\documentclass{article}
\usepackage[hyphens]{url}

\usepackage{microtype}
\usepackage{graphicx}
\usepackage{subfigure}
\usepackage{booktabs} 

\usepackage{hyperref}

\usepackage[dvipsnames]{xcolor}

\usepackage[accepted]{icml2024}

\usepackage{amsmath}
\usepackage{amssymb}
\usepackage{mathtools}
\usepackage{amsthm}
\usepackage{dsfont}

\usepackage[capitalize,noabbrev]{cleveref}

\theoremstyle{plain}
\newtheorem{theorem}{Theorem}[section]

\newtheorem{lemma}[theorem]{Lemma}

\theoremstyle{definition}

\theoremstyle{remark}

\usepackage[textsize=tiny]{todonotes}

\usepackage{ulem}
\usepackage{graphicx}
\usepackage{booktabs} 
\usepackage{tabularx}
\usepackage{textcomp}
\usepackage{bm}

\renewcommand{\emph}[1]{\textit{#1}}

\usepackage{amsmath}
\usepackage{amsfonts}
\usepackage{colortbl}

\usepackage{listings}
\usepackage{multicol}
\usepackage{multirow}
\definecolor{codegreen}{rgb}{0,0.6,0}
\definecolor{codegray}{rgb}{0.5,0.5,0.5}
\definecolor{codepurple}{rgb}{0.58,0,0.82}
\definecolor{backcolour}{rgb}{0.95,0.95,0.92}

\lstdefinestyle{mystyle}{
  backgroundcolor=\color{backcolour}, commentstyle=\color{codegreen},
  keywordstyle=\color{magenta},
  numberstyle=\tiny\color{codegray},
  stringstyle=\color{codepurple},
  basicstyle=\ttfamily\footnotesize,
  breakatwhitespace=false,         
  breaklines=true,                 
  captionpos=b,                    
  keepspaces=true,                 
  numbers=left,
  xleftmargin=2em,
  frame=single,
  framexleftmargin=1.5em,
  numbersep=5pt,                  
  showspaces=false,                
  showstringspaces=false,
  showtabs=false,                  
  tabsize=2,
  float
}

\icmltitlerunning{Dissecting Multimodality in VideoQA Transformer Models by Impairing Modality Fusion}

\begin{document}

\twocolumn[
\icmltitle{{D}issecting {M}ultimodality in {V}ideo{QA} {T}ransformer {M}odels \\ by {I}mpairing {M}odality {F}usion}




\begin{icmlauthorlist}
\icmlauthor{Ishaan Singh Rawal}{ihpc,cfar}
\icmlauthor{Alexander Matyasko}{ihpc,cfar}
\icmlauthor{Shantanu Jaiswal}{ihpc,cfar}
\icmlauthor{Basura Fernando}{ihpc,cfar}
\icmlauthor{Cheston Tan}{ihpc,cfar,ir}
\end{icmlauthorlist}

\icmlaffiliation{cfar}{
Institute of High Performance Computing, Agency for Science, Technology \& Research, Singapore}
\icmlaffiliation{ihpc}{Centre for Frontier AI Research, Agency for Science, Technology \& Research, Singapore}
\icmlaffiliation{ir}{Institute for Infocomm Research, Agency for Science, Technology \& Research, Singapore}

\icmlcorrespondingauthor{Ishaan Singh Rawal}{rawal\_ishaan\_singh@cfar.a-star.edu.sg}

\icmlkeywords{Machine Learning, ICML, Deep Learning, Multimodal Learning, explainable AI, xAI}

\vskip 0.3in]



\printAffiliationsAndNotice{}  

\begin{abstract}
While VideoQA Transformer models demonstrate competitive performance on standard benchmarks, the reasons behind their success are not fully understood. Do these models capture the rich multimodal structures and dynamics from video and text jointly? Or are they achieving high scores by exploiting biases and spurious features? Hence, to provide insights, we design \textit{QUAG} (QUadrant AveraGe), a lightweight and non-parametric probe, to conduct dataset-model combined representation analysis by impairing modality fusion. We find that the models achieve high performance on many datasets without leveraging multimodal representations. To validate QUAG further, we design \textit{QUAG-attention}, a less-expressive replacement of self-attention with restricted token interactions. Models with QUAG-attention achieve similar performance with significantly fewer multiplication operations without any finetuning. Our findings raise doubts about the current models' abilities to learn highly-coupled multimodal representations. Hence, we design the \textit{CLAVI} (Complements in LAnguage and VIdeo) dataset, a stress-test dataset curated by augmenting real-world videos to have high modality coupling. Consistent with the findings of QUAG, we find that most of the models achieve near-trivial performance on CLAVI. This reasserts the limitations of current models for learning highly-coupled multimodal representations, that is
not evaluated by the current datasets (project page: \url{https://dissect-videoqa.github.io}).


\end{abstract}
\section{Introduction}
Multimodal learning with videos and language presents a challenge, given their shared sequential nature but distinct underlying structures. That is, videos exhibit spatio-temporal dynamics in the pixel space, whereas language representation is composed of the syntax and semantics of word sequences. Hence, tasks like Video Question Answering (VideoQA) \citep{zhong2022video} are difficult as they necessitate the model to acquire accurate representations of both modalities and establish meaningful connections between them. Transformers have demonstrated exceptional performance on VideoQA benchmarks \citep{zhong2022video}. But does the good performance of Transformers on current VideoQA benchmarks necessarily mean that they learn to faithfully represent, leverage, understand, and reason about the modalities? Or do the current benchmarks and metrics fail to robustly evaluate the models for their multimodal understanding? 

This is a valid concern because deep learning models can learn shortcuts to achieve good performance without leveraging and aligning the modalities \citep{geirhos2020shortcut}. For example, seemingly spatio-temporal tasks, like some action classification problems, are shown to be solved without focusing much on temporal representations \citep{sevilla2021only, kowal2022deeper}. Similarly, in VideoQA, recent works report that the datasets contain specific biases \citep{buch2022revisiting, lei2022revealing}. However, most studies on biases are restricted to isolated analyses of either the models or the datasets. This raises question: \textbf{Do VideoQA models learn to jointly leverage the modalities to achieve competitive performance on current benchmarks?}

To answer these questions, we propose QUAG 
(QUadrant AveraGe), a lightweight and non-parametric probe to systematically gauge the reliance of a finetuned model's performance on joint multimodal representations.  QUAG impairs modality fusion by block-averaging attention weights. We apply QUAG on multiple dataset-model combinations, and consistently find that the models manage to achieve high performance on the benchmarks without leveraging multimodal representations. This finding is concerning because high performance on established benchmarks should be ideally indicative of coupled multimodal understanding. We confirm this by achieving similar performance by replacing self-attention with QUAG-attention, a restricted variant of self-attention, without any finetuning. This might be because the models might be leveraging the spurious biases in the datasets (shortcuts), that weaken the modality coupling (we refer the degree of interdependence or interaction between different  modalities as {modality coupling} \citep{XIE20072325, hu2022voxel, liu2023modality}. Then, \textbf{how do VideoQA models perform in highly-coupled multimodal settings?}

Thus, we create CLAVI (Complements in LAnguage and VIsion), a stress-test dataset with high multimodal coupling. It is  created automatically by augmenting real-world videos and template-generated temporal questions (Figure \ref{fig:clavi}). We observe that nearly all fine-tuned models exhibit high accuracy on shortcut instances (that is, lower multimodal coupling) in CLAVI, but most have near-trivial performance on highly-coupled multimodal instances. This reasserts the limitations of current models for learning highly-coupled multimodal representations, that is not tested in the current benchmarks.

In summary, our contributions are \emph{(i)}  We design QUAG, a systematic method for assessing the relative contribution of various multimodal components in the modality fusion stage; \emph{(ii)} Using QUAG and QUAG-attention, we demonstrate that high performance on established VideoQA benchmarks is not completely representative of coupled multimodal understanding; and \emph{(iii)} We develop CLAVI, a new stress-test for VideoQA with high multimodal coupling. Overall, QUAG and CLAVI demonstrate that the current VideoQA models are not proficient at learning highly-coupled multimodal representations, a crucial aspect that is not systematically tested in the current benchmarks.
\section{Do VideoQA Models Learn to Jointly Leverage the Modalities?}
\label{methods}

\graphicspath{ {./images} }

\newcommand{\red}[1]{%
  \colorbox{red!15}{$\displaystyle#1$}}
\newcommand{\blue}[1]{%
  \colorbox{blue!15}{$\displaystyle#1$}}
\newcommand{\yellow}[1]{%
  \colorbox{yellow!15}{$\displaystyle#1$}}

\newcommand{\hred}[2][red!15]{\mathpalette{\highlightwithstyle[#1]}{#2}}
\newcommand{\hblue}[2][blue!15]{\mathpalette{\highlightwithstyle[#1]}{#2}}
\newcommand{\hyellow}[2][yellow!15]{\mathpalette{\highlightwithstyle[#1]}{#2}}

\newcommand{\highlightwithstyle}[3][red!15]{
  \begingroup                         
    \sbox0{$\mathsurround 0pt #2#3$}
    \setlength{\fboxsep}{.5pt}        
    \sbox2{\hspace{-.5pt}
      \colorbox{#1}{\usebox0}
    }%
    \dp2=\dp0 \ht2=\ht0 \wd2=\wd0     
    \box2                             
  \endgroup                           
}

\newlength{\Width}%

Transformer architecture has become the de facto model for solving the VideoQA task in recent years. However, since they lack the relevant intrinsic inductive biases, they must learn it from the data \citep{xu2021vitae}. For an ideal multimodal model and dataset, high accuracy should be achievable only by aligning and leveraging both -- the text (question) and the visual (video) modalities. Despite this,  datasets often contain biases that may not necessitate multimodal understanding. However, the presence of spurious features in a dataset does not always imply that the model is exploiting them \cite{murali2023shortcut}. Hence, we introduce QUAG for combined dataset-model representation analysis. 

In the context of multimodal Transformers, we posit that modality fusion layers enable multimodal understanding by progressively attending to informative tokens within and between modalities. QUAG systematically ablates the effects of multimodal attention by augmenting self-attention in the fusion layer. 
For instance, for a given dataset, if the model heavily relies only on unimodal information, ablating the unimodal component of modality fusion should significantly decrease performance. Also, this decrease in performance should be more pronounced than the decrease from ablating the crossmodal component of attention.
QUAG achieves such specific modality interaction ablation through block averaging, as explained below.

\subsection{Video Question Answering Setup} 
In VideoQA, the task is to predict the correct answer given a video-question tuple, $(\mathcal{V}, \mathcal{T})$. A VideoQA model consists of a vision encoder $ F_\mathcal{V}: \mathcal{V} \rightarrow \mathbb{R}^{l_\mathcal{V} \times d}$, text encoder $ F_\mathcal{T}: \mathcal{T} \rightarrow \mathbb{R}^{l_\mathcal{T} \times d}$, a multimodal fusion module $M: ( F_\mathcal{V}(\mathcal{V}),  F_\mathcal{T}(\mathcal{T}))\rightarrow \mathbb{R}^{ {(l_\mathcal{V}+l_\mathcal{T})} \times d}$, and a classification  layer, where $l_\mathcal{V}$ and $l_\mathcal{T}$ are the maximum input sequence lengths of video and text modalities respectively and $d$ is the dimensionality of the fusion model.

Consider $M$ as a composition of $n$ attention-based multimodal fusion blocks, $M = M_n \circ M_{n-1} \circ \cdots M_1$. Each fusion block consists of attention, normalization, and token-mixing modules. For our analysis, we consider $M$ to be composed of self-attention transformer blocks. That is,  query, key, and value are the transformations of the same input sequence. $ \bm X_{\mathcal{V}\mathcal{T}} = [ F_\mathcal{V}(\mathcal{V}) \mathbin\Vert  F_\mathcal{T}(\mathcal{T})] \in \mathbb{R}^{(l_\mathcal{V} + l_\mathcal{T}) \times d}$ is the input for $M$, where $\mathbin\Vert$ is concatenation operator for the token dimension. Since QUAG is applied at inference time only, we assume the VideoQA model to be frozen.

\subsection{QUAG: Ablation of Modality Interactions}

Shortcuts are the spurious features learned by a given model on a given dataset \citep{murali2023shortcut}. Along this axis, we use QUAG to pinpoint the exact failure modes in the datasets representations learned by the models. 

Let $ \bm{X}_{i-1}$ denote the input of the fusion block $M_i$ and let $( \bm{Q}_i,  \bm{K}_i,  \bm{V}_i)$ be its query, key, and value transformations and $\bm X_0 =  \bm X_{\mathcal{V}\mathcal{T}}$. Then, the token-mixing operation is given by $ \bm T_i =  \bm A_i  \bm V_i$, where $ \bm A_i = \sigma ( \bm{Q}_i  \bm{K}_i^\top)$ is the attention matrix (we denote the softmax operation by $\sigma$ and omit the scaling factor $\sqrt{d}$ for readability). For ${ \bm{Q}_{1u}}$, ${ \bm{K}_{1u}}$, and ${ \bm{V}_{1u}}$ to denote the query, key, and value projections of modality $u$ for the first fusion block, $M_1$,  we can simplify, $ \bm{A}_1$ and $ \bm{T}_1$ in terms of their partition blocks, referred to as quadrants henceforth, as: 
\[
      \bm{\bm{A}}_1 =
    \sigma \left(\left[
    \begin{array}{c|c}
    [\red{\bm{ Q}_{1\mathcal{\bm{V}}}^{\vphantom{\top}}} \red{ \bm{K}^\top_{1\mathcal{\bm{V}}}}]_{_{l_\mathcal{\bm{V}} \times l_\mathcal{\bm{V}}}}  
    & [\red{{\bm Q}_{1\mathcal{\bm{V}}}^{\vphantom{\top}}} 
    \blue{ \bm{K}^\top_{1\mathcal{T}} }]_{_{l_\mathcal{\bm{V}} \times l_\mathcal{T}}} \\[5pt]
    \hline \\[-5pt]
    [\blue{ \bm{Q}_{1\mathcal{T}}^{\vphantom{\top}}} \red{ \bm{K}^\top_{1\mathcal{\bm{V}}}}]_{_{l_\mathcal{T} \times l_\mathcal{\bm{V}}}}  & [\blue{ \bm{Q}_{1\mathcal{T}}^{\vphantom{\top}}} \blue{ \bm{K}^\top_{1\mathcal{T}}}]_{_{l_\mathcal{T} \times l_\mathcal{T}}}
    \end{array} \right] \right) 
\]
\[
     \bm{T}_1 =
    \left[
    \begin{array}{c|c}
    { \bm{A}^1_{\hred{\mathcal{\bm{V}}}\hred{\mathcal{\bm{V}}}}} &   \bm{A}^1_{\hred{\mathcal{\bm{V}}}\hblue{\mathcal{T}}} \\[5pt]
    \hline \\[-5pt]
    { \bm{A}^1_{\hblue{\mathcal{T}}\hred{\mathcal{\bm{V}}}}} &   \bm{A}^1_{\hblue{\mathcal{T}}\hblue{\mathcal{T}}}
    \end{array} \right]
    \left[
    \begin{array}{c}
     \bm{V}_{1\hred{\mathcal{\bm{V}}}}^{\vphantom{\top}} \\[5pt]
    \hline \\[-5pt]
     \bm{V}_{1\hblue{\mathcal{T}}}^{\vphantom{\top}}
    \end{array} \right] 
\]

where $ \bm{A}^1_{{u_1}{u_2}}$ represents the quadrant of $ \bm{A}_1$ corresponding to (${ \bm{Q}_{1{u_1}}^{\vphantom{\top}}} { \bm K^\top_{1{u_2}}}$).  Note that we skip layer normalization in the discussion for simplicity. We can simplify $ \bm T_1$ as: 
\begin{equation}
    \label{eq:t1}
     \bm T_1 =
    \left[
    \begin{array}{c}
    { \bm A^1_{\hred{\mathcal{V}}\hred{\mathcal{V}}}}  \bm V_{1\hred{\mathcal{V}}}^{\vphantom{\top}} +  \bm A^1_{\hred{\mathcal{V}}\hblue{\mathcal{T}}}  \bm V_{1\hblue{\mathcal{T}}}^{\vphantom{\top}} \\[5pt]
    \hline \\[-5pt]
    { \bm A^1_{\hblue{\mathcal{T}}\hred{\mathcal{V}}}}  \bm V_{1\hred{\mathcal{V}}}^{\vphantom{\top}} +  \bm A^1_{\hblue{\mathcal{T}}\hblue{\mathcal{T}}}  \bm V_{1\hblue{\mathcal{T}}}^{\vphantom{\top}}
    \end{array} \right] 
\end{equation}

Following the quadrant partitioning scheme of $\bm A_1$ in $\bm M_1$, we define quadrants for all the attention matrices in the downstream fusion layers.

Next, we define row-wise average-and-replace operator $\mathcal{R}$ that operates on a quadrant. $\mathcal{R}$ replaces all the values in a given partitioned row with the respective mean value.  Note that the values in the other quadrants are unaffected. 
Given a matrix $ \bm{Z} \textrm{ of size } p \times q$ and let $W$ denote the location of the quadrant of $ \bm{Z}$ with indices $(p_1^W \cdots p_2^W) \times (q_1^W \cdots q_2^W)$. We use $[~ . ~]_{ij}$ to index the element in row $i$ and column $j$. Then, $\mathcal{R}$ is formally defined as:
\begin{equation*}
    \displaystyle [\mathcal{R}( \bm{Z},W)]_{ij} \coloneqq \begin{cases} 
          \sum_{k=q^W_1}^{q^W_2} \frac{[{\bm Z}]_{ik}}{q^W_2 - q^W_1 + 1}& 
          \begin{aligned}
          i \in \{p_1^W, \cdots p_2^W\} \\ j \in \{q_1^W, \cdots q_2^W\}
          \end{aligned}
          \vspace{-1em}
          \\
          \\ 
          [\bm{ Z}]_{ij} & \text{otherwise} 
       \end{cases}  
\end{equation*}

\begin{figure}
  \centering
  \includegraphics[scale=0.23]{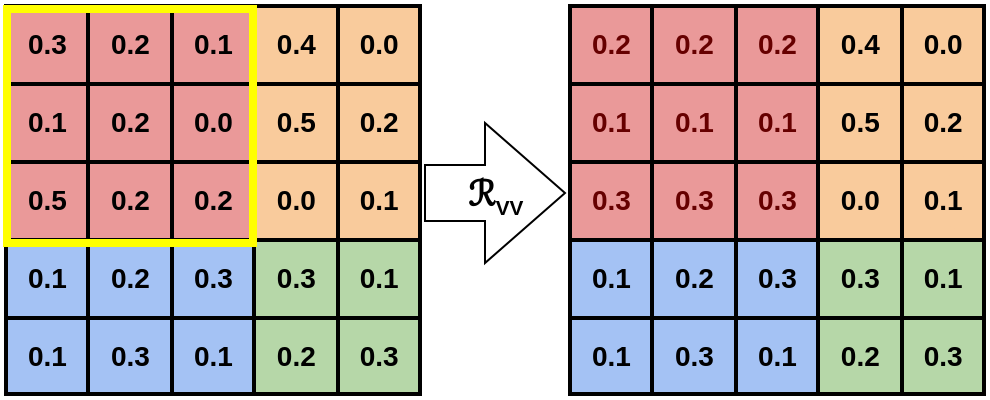}
  \caption{Illustrative toy example of row-wise average-and-replace operation or $\mathcal{R}( \bm{A}, \mathcal{VV})$, where $ \bm{A}$ is the input attention matrix (left). The cells are colored as per their quadrants ($\mathcal{VV: \textrm{red}}, \mathcal{VT: \textrm{yellow}}, \mathcal{TV: \textrm{blue}}, \mathcal{TT: \textrm{green}}$).  We apply $\mathcal{R}$ operator on the $\mathcal{VV}$ quadrant (highlighted in yellow)  to replace the values with the respective row-wise average value (right).}
  \label{fig:main_quag_example}
\end{figure}

We can now formally define the QUAG operator, $\phi$, as: 
\[
\phi( \bm A_i, [s_1, s_2, \cdots, s_m]) \coloneqq (\mathcal{R}_{s_1} \circ \mathcal{R}_{s_2} \cdots \circ \mathcal{R}_{s_m} ( \bm A_i)), 
\]

where {$S = [s_1, s_2, \cdots, s_m]$ is a list of quadrants such that} ${\forall s \in S: } s \in \{\mathcal{T}\mathcal{T}, \mathcal{T}\mathcal{V}, \mathcal{V}\mathcal{T}, \mathcal{V}\mathcal{V}\}$, $\mathcal{R}_{s_i}( A_i)$ is short-hand for $\mathcal{R}( \bm A_i,s_i)$, $ \bm A_i$ is the attention matrix of fusion layer $M_i$. Note that $\mathcal{T}\mathcal{T}$ refers to the quadrant $\bm A^i_{\mathcal{T}\mathcal{T}}$ of $\bm A_i$ and so on. Refer Figure \ref{fig:main_quag_example} for an illustrative example. We define QUAG operation on a layer $i$ as a modification of the existing attention matrix, that is, $
 \bm A_i \leftarrow \phi( \bm A_i, [s_1, s_2, \cdots, s_m]) $.  Incorporating QUAG into the existing model pipeline is straightforward and we provide the code in the Appendix \ref{sec:quag_pseudo_code}. Since we will be applying the QUAG operator successively on all the layers of $M$, for brevity, we denote $\Phi(M,n,S) = \forall_{i \in [1, \cdots, n]} \; \bm A_i \leftarrow \phi( \bm A_i, S)$. Note that QUAG is \textbf{light-weight}, \textbf{non-parametric}, requires \textbf{no finetuning} and operates at \textbf{inference time} for combined dataset-model analysis.
\subsection{Short-circuit Operations}

As QUAG is a generic method, we consider some special cases, based on the value of $S$, below. We call these operations collectively as short-circuiting:

    1) $\bm{S=[\mathcal{V}\mathcal{V}, \mathcal{T}\mathcal{T}]}$: As evident from Eqn. \ref{eq:t1}, in the token-mixing step, the QUAG operation $\bm{A}_1 \leftarrow \phi( \bm{A}_1, [\mathcal{V}\mathcal{V}, \mathcal{T}\mathcal{T}])$ results in scaling the average values of $ \bm{V}_{1\mathcal{V}}$ and $ \bm{V}_{1\mathcal{T}}$ in the upper and lower blocks of $ T_1$ respectively. Mathematically, we can represent it as $\phi(\bm{A}_1, [\mathcal{V}\mathcal{V}, \mathcal{T}\mathcal{T}]){_{\mathcal{V} \mathcal{V}}} \bm{V}{_{1\mathcal{V}}} = \bm{\mu}_{{\mathcal{V} \mathcal{V}}} \bm{1} \bm{V}_{1\mathcal{V}}$, where $\bm{\mu}_{{\mathcal{V} \mathcal{V}}}$ is a column vector containing row-wise average values of $\bm{A}_{\mathcal{V}\mathcal{V}}$ and $\bm{1}$ is a row vector of ones. Clearly, $\bm{1}\bm{V}{_{1\mathcal{V}}}$ is equivalent to averaging across the video features. Similarly, $\mathcal{TT}$ features are averaged in the lower block. Note that the crossmodal attention weights, $\bm{A}^1_{\mathcal{V}\mathcal{T}}$ and $\bm{A}^1_{\mathcal{T}\mathcal{V}}$ are unchanged, and only the unimodal token mixing is \textit{short-circuited}. We name such a fusion block as unimodal average conformable. And when applied to all the fusion blocks, the operation is known as \textbf{unimodal short-circuiting}. We provide a toy example in \S \ref{sec:toy_example_sc}.

    \begin{theorem}
    \label{theorem:unimodal_sc}
        {Unimodal short-circuiting produces unimodal average conformable features in all the fusion blocks.}
    \end{theorem}
    \begin{proof}
    \label{proof:unimodal_sc}
        We use mathematical induction to prove that $\Phi(M, n, [\mathcal{V}\mathcal{V}, \mathcal{T}\mathcal{T}])$ makes all the $n$ fusion blocks in $M$  unimodal average conformable. 
        
        We have already explained the base case above. 
        
        Let us assume that the induction hypothesis is true for first $L-1$ fusion blocks in $M$. That is, given that $\Phi(M, L-1, [\mathcal{VV}, \mathcal{TT}])$ is unimodal average conformable, we want to show that $\bm{A}_L \leftarrow \phi(\bm{A}_{L}, [\mathcal{VV},\mathcal{TT}])$ will make the output of $M$ unimodal average conformable. 
        
        Because of skip connections in transformers, $\bm{X}_L$ (the input of $M_L$) can be decomposed as a function of $(\bm{X}_{L-1} + \bm{T}_{L-1})$. Note that while the function is non-linear, it is applied point-wise. That is, there is no token mixing and hence, the conformability is not affected. By the inductive hypothesis, we know that $\bm{T}_{L-1}$ is unimodal average conformable. We can recursively decompose $\bm{X}_{L-1}$ and show that the only unconformable component is the input, $\bm{X}_0$. As shown in the base case, $\phi$ operated on a fusion block with input $\bm{X}_0$ makes it unimodal average conformable. Hence, given $\Phi(M, L-1, [\mathcal{VV}, \mathcal{TT}])$ is unimodal average conformable, $\bm{A}_L \leftarrow \phi(\bm{A}_{L}, [\mathcal{VV},\mathcal{TT}])$ will make $M_L$ unimodal average conformable. Hence, overall,  $\Phi(M, n, [\mathcal{VV}, \mathcal{TT}])$ makes $M$ unimodal average conformable. 
    \end{proof}
    
    Effectively, $\Phi(M,n,[\mathcal{V}\mathcal{V}, \mathcal{T}\mathcal{T}])$ bypasses the effect of video-video attention and text-text attention. We proved above that unimodal token-mixing is reduced to scaling the average of the modalities.  That is, it ablates unimodal representations to analyze their dependence on the performance of the models.
    Since the following cases can be proved similarly using induction, we skip the proofs for conciseness. 

    2) $\bm{S =  [\mathcal{V}\mathcal{T}, \mathcal{T}\mathcal{V}]}$: Parallel to unimodal short-circuiting, $\phi( \bm{A}_1, [\mathcal{V}\mathcal{T}, \mathcal{T}\mathcal{V}])$ is equivalent to scaling the average values of $ \bm{V}_{1\mathcal{T}}$ and $ \bm{V}_{1\mathcal{V}}$ in the upper and lower blocks of $ \bm{T}_1$ respectively. While the crossmodal attention in video-text is reduced to uniform attention, the unimodal components of attention are unaffected. We term this effect as \textbf{crossmodal short-circuiting}. It is complementary to unimodal short-circuiting and assesses the importance of inter-modality token-mixing. It probes if the models actually learns by fusing the information between the two modalities or is it largely driven by unimodal information.
    
    3) $\bm{S = [\mathcal{V}\mathcal{V}, \mathcal{T}\mathcal{V}]}$: This is equivalent to ablating the effect of individual of video keys, resulting in averaging the components of video modality in the upper and lower blocks of all $ \bm{T}_i$. We call this \textbf{video short-circuiting}. 
    Similarly, $\bm{S = [\mathcal{T}\mathcal{T}, \mathcal{V}\mathcal{T}]}$ leads to \textbf{text short-circuiting}.  Singular values and their properties have been used to study the information richness of attention matrices. For example, \citet{gui-xiao-2023-hifi} use top-k singular values to measure the richness of attention matrix. We prove that video and text short-circuiting reduces the upper bound on the rank (the number of non-zero singular values) of the attention matrix. This in turn reduces the maximum representation power of the attention operation \citep{bhojanapalli2020low}.
    \begin{lemma}
    \label{lem:rank_reduction}
    For an attention matrix $\bm A$ of size $(l_\mathcal{V} + l_\mathcal{T}) \times (l_\mathcal{V} + l_\mathcal{T})$, the rank, $\rho$, after video short-circuiting is bounded as: $\rho(\phi(\bm A, [\mathcal{VV, TV}])) \leq l_\mathcal{T} + 2$.
    \end{lemma}
    We provide the proof in the \S \ref{sec:lemma_proof}. Note that the similar bound exists for text short-circuiting.


\subsection{QUAG-attention}
 Along with the assessment of multimodal understanding, QUAG enables a detailed analysis of token mixing for identifying the sub-optimality of learned representations. Sub-optimality occurs if the modality fusion process doesn't effectively capture the information within each modality along with the complementary information in the other modality, even with computationally expensive operations like self-attention. Hence, we propose QUAG-attention to rely only on the sub-optimal representations learnt by the model by restricted token mixing, without finetuning.
 
  QUAG-attention, inspired by text and video short-circuiting operations, is a replacement of self-attention in fusion module that performs attention calculation and token-mixing on already short-circuited sequences. That is, QUAG-attention calculates attention using already averaged key tokens and applies on averaged value tokens, thus effectively pruning the number of keys and values (as opposed to block-averaging post attention calculation in QUAG).

  As proved in \cref{theorem:unimodal_sc}, consistent averaging of partitions maintains their average conformability. Extending this idea for QUAG-attention, we average the partition block corresponding to the modality in each of the fusion layers before transforming the tokens into keys and values. For example, if the input of the model is $\bm{X}_{\mathcal{V}\mathcal{T}} = [ F_\mathcal{V}({\mathcal{V}}) \mathbin\Vert  F_\mathcal{T}({\mathcal{T}})]$, then consistently averaging the upper partition block (corresponding to the video modality), before key and value transformation would result in video-average QUAG-attention. QUAG-attention can be applied to only text, video or both the modalities. It reduces the number of key and value tokens from $(l_\mathcal{V} + l_\mathcal{T})$ to either $(l_\mathcal{T}+1)$ (text-average), $(l_\mathcal{V} + 1)$ (video-average) or $2$ (text-video-average) in all the multimodal fusion blocks while the number of query tokens remain the same. Each averaged token in QUAG-attention is representative of multiple tokens. This might cause the average tokens to have less contribution in the output of softmax operation. Hence, we use scale the softmax function (similar to proportional attention \citep{bolya2022tome} as $\bm{A} = \sigma (\frac{\bm{QK}^\top}{\sqrt{d}} + log(\bm{S}))$, where $\bm{S}$ is a row vector containing the effective size of each token (that is, the number of tokens  the average token is representing)).

  Similar to QUAG, QUAG-attention also require no finetuning or additional parameters. High accuracy  with QUAG-average is a concern because that would mean that the model never learnt to leverage that modality effectively, yet manages to achieve high scores on the multimodal benchmark.

\section{Simulation Study}
\label{sec:simulation_study}
The amount of modality coupling affects the kind of representations the model learns.  We want to analyze the effects of degree of coupling between modalities on the learnt representations using QUAG. However, accurately characterizing the distribution of multimodal data is an intractable problem. Therefore, it is not possible to quantify the degree of coupling between the modalities. Hence, inspired by \citet{huang2021makes}, we propose a simple method of generating multimodal data with known degree of coupling.

\textbf{Data Generation: } Consider two modalities -- $t$ and $v$ with dimensionality 100. Each sample contains 15 tokens from each modality. Below is the sample generation process:

\textbf{Step 1:} Generate $m_i \sim \mathcal{N}(\bm 0, \textrm{\textbf{I}})$, for $i=1,2$, where $m_i$ is of size $[15 \times 100]$. Let $z = [m_1 \mathbin\Vert  m_2]$.

\textbf{Step 2:} Generate the 30-dimensional output vector $y$ by taking weighted mean across the feature dimension. So, $[y]_{i} = \frac{1}{100}\sum^{100}_{j=1} {p_{j} \cdot [z]_{ij}}$, where $p_j = 10 \cdot \sin{\frac{j \pi}{202}}$ and $i = 1, 2, \cdots , 30$. We choose weighted mean over mean, because the mean value tends to zero.

\textbf{Step 3:} Generate $t$ and $v$. We  define the following transformations: $t \leftarrow (1-\alpha) \cdot m_1 - \alpha \cdot m_2$ and $v \leftarrow (1-\alpha) \cdot m_2 - \alpha \cdot m_1$, where $\alpha \in (0, 0.5)$

Where $\alpha$ is the crossmodal coupling coefficient. Increasing $\alpha$ increases the degree of crossmodal token mixing in the inputs. We  use \textbf{modality coupling} to jointly refer to unimodal and crossmodal coupling in the proceeding sections. For simplicity we consider cross-modality coupling to be linear but it might be more complex for real data.

\begin{figure}[h!]
  \centering
  \includegraphics[scale=0.44]{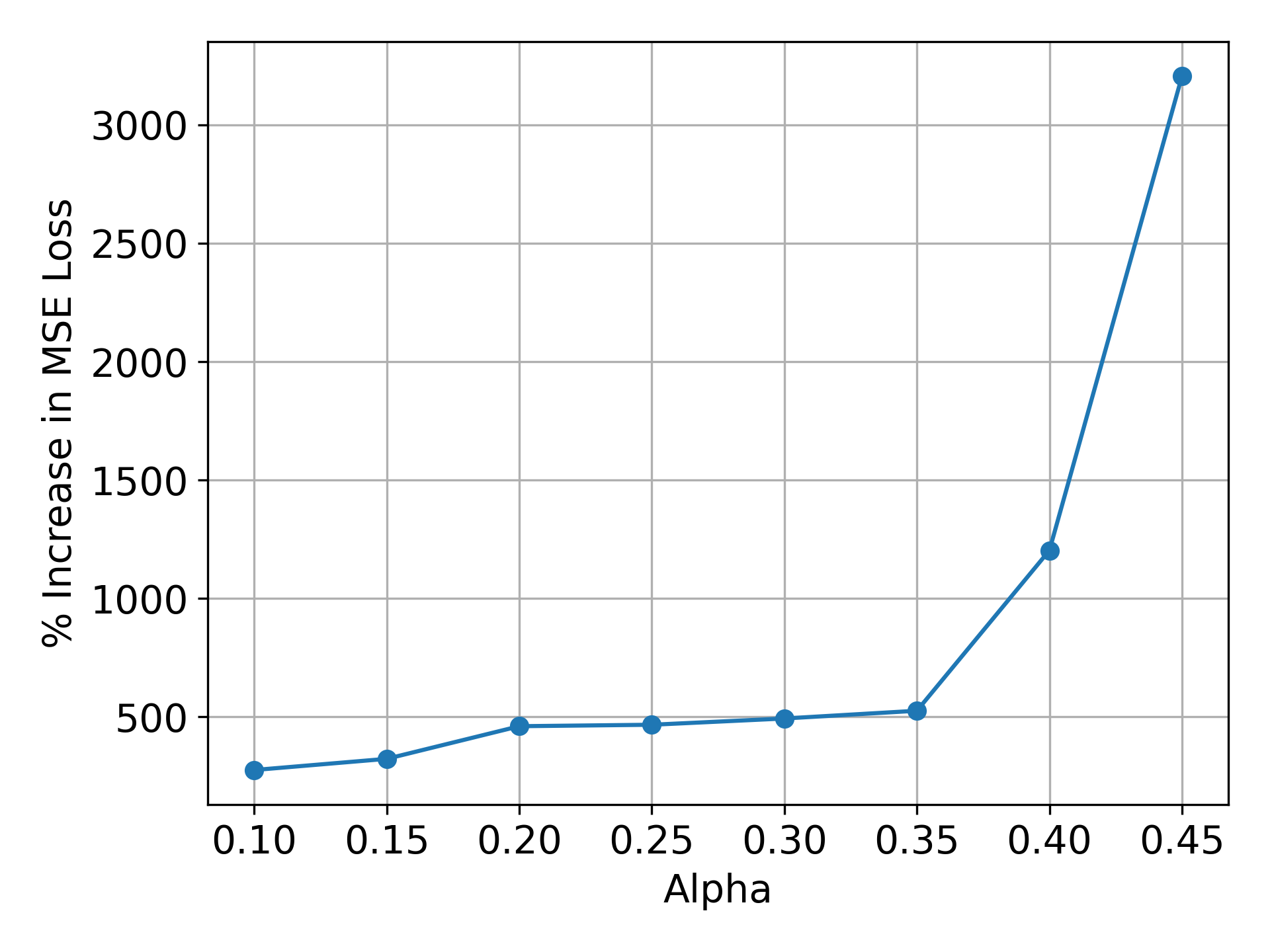}
  \caption{Result of the simulation study. The plot of percentage increase in the mean squared loss after crossmodal short-circuiting (y) versus $\alpha$, the crossmodal coupling coefficient (x). }
  \label{fig:quag_sim}
\end{figure}

We train a simple transformer model on the crossmodal data for different values of $\alpha$. The model and training details are provided in \S \ref{sec:supp_sim}. As shown in \cref{fig:quag_sim},  we find that as $\alpha$ increases, the test error after crossmodal short-circuiting also increases. This is expected because increasing $\alpha$ would necessitate the model to rely increasingly on crossmodal information. Therefore, ablating crossmodal interactions with QUAG has increasingly deleterious effect. \textbf{This validates QUAG as a method of impairing multimodal fusion}. Using this understanding, in the next section, we apply QUAG to the models trained on real-world datasets to analyze the modality interactions in the model.

\section{Combined Dataset-Model Analysis}
\label{exp_set}
 Ideally, for optimal multimodal representations learnt on a dataset, we expect dramatic drops in all the short-circuiting operations. However, datasets contain biases and the model may or may not learn them. Hence, we perform combined dataset-model analysis with QUAG to find the reliance of a model on the biases of a dataset through performance drops under short-circuiting operations.
 

We evaluate QUAG and QUAG-attention on JustAsk \citep{yang2021justask} and FrozenBiLM  \citep{yang2022zero} models. We evalaute it on the following datasets  ActivityNet-QA \citep{yu2019activitynet}, {MSRVTT-QA} \citep{xu2017video}, and {NeXT-QA} \citep{xiao2021next} We also report results on the {ATP-Hard} subset of NeXT-QA \citep{buch2022revisiting} that contains a higher concentration of temporally challenging data requiring multi-frame understanding. We provide the implementation details in  \S \ref{sec:quag_impl_detail}

\subsection{Analysis}

\begin{table*}
  \caption{Short-circuit (SC) and QUAG-attention accuracies for JustAsk and FrozenBiLM models on ActivityNet-QA (A-QA), MSRVTT-QA (M-QA), NeXT-QA (N-QA) and ATP-Hard (ATP-H) datasets (*: video-average for FrozenBiLM and video-text-average for JustAsk; ${}^{\dag}$: \% decrease in multiplication operations due to QUAG-attention; $\downarrow$: drop $\geq$ 2\% with respect to baseline (the higher drop, the better). }

  \label{quag-res}
  \centering
  \begin{tabular}{llllllllll}
    \toprule
    \multicolumn{1}{c}{} &
    \multicolumn{4}{c }{FrozenBiLM} &
    \multicolumn{1}{c}{} &
    \multicolumn{4}{c}{JustAsk} \\
    \cmidrule(r){2-5}
    \cmidrule(r){7-10}
        ~ & A-QA & M-QA & N-QA & ATP-H &  & A-QA & M-QA & N-QA & ATP-H \\ 
        \midrule 
        Baseline & 43.6 & 46.6 & 55.8 & 55.7 &  & 38.7 & 41.8 & 53.8 & 44.0 \\
        Language-only & 32.2 $\downarrow$ & 33.2 $\downarrow$ & 55.7 & 55.8 &  & 28.2 $\downarrow$ & 29.9 $\downarrow$ & 42.2 $\downarrow$ & 42.0 \\ 
        Video-only & {\hphantom{2}}0.1 $\downarrow$ & {\hphantom{2}}0.0 $\downarrow$ & 20.2 $\downarrow$ & 20.1 $\downarrow$ &  & {\hphantom{2}}2.6 $\downarrow$ & {\hphantom{2}}6.7 $\downarrow$ & 39.1 $\downarrow$ & 23.0 $\downarrow$ \\
        \midrule
        SC: unimodal & {\hphantom{2}}2.4 $\downarrow$ & {\hphantom{2}}1.0 $\downarrow$ & 19.8 $\downarrow$ & 21.4 $\downarrow$ &  & 38.5 & 41.5 & 53.6 & 43.6 \\ 
        SC: crossmodal & 32.3 $\downarrow$ & 32.8 $\downarrow$ & 56.0 & 55.6 &  & 38.3 & 41.3 & 53.5 & 44.3 \\ 
        SC: video & 43.1 & 45.7 & 55.8 & 55.7 &  & 38.2 & 41.3 & 53.4 & 44.3 \\ 
        SC: text & {\hphantom{2}}1.4 $\downarrow$ & {\hphantom{2}}1.0 $\downarrow$ & 20.5 $\downarrow$ & 21.1 $\downarrow$ &  & 38.6 & 41.5 & 53.7 & 43.6 \\
        \midrule
        QUAG-atten* & 43.0 & 45.8 & 55.6 & 55.9 &  & 38.0 & 41.0 & 53.5 & 44.1 \\ 
    {$\Delta$Multiplication Ops${}^\dag$} & \multicolumn{4}{c}{\bf{13.6\%}} & \multicolumn{5}{c}{\bf{68.0\%}} \\
    \bottomrule
  \end{tabular}
\end{table*}




The results are shown in Table \ref{quag-res}. For comparison to the unperturbed model, we specify the baseline, language-only (without video input) and video-only (without text input) accuracies. The high performance in language-only setting relative to the baseline is indicative of strong unimodal bias towards language. However, these metrics do not provide any information about the exact nature and degree of the sub-optimal representations learned by the models.

\textbf{(1)} The performance of FrozenBiLM on ActivityNet-QA and MSRVTT-QA drops by over 10\%  with crossmodal short-circuiting, and by 40\% with both unimodal and text short-circuiting. Furthermore, the drop is less than 1\% under video short-circuiting.  
However, for NeXT-QA and ATP-Hard, the performance of FrozenBiLM drops to chance level (20\%) under text and unimodal short-circuiting operations but hardly drops with video and text short-circuiting. This means that FrozenBiLM consistently has strong reliance unimodal interactions and the text modality. However, the model learns to leverage crossmodal representations only for ActivityNet-QA and MSRVTT-QA and not NeXT-QA and ATP-Hard. 
\\
\textbf{(2)} On the other hand, the performance of JustAsk model does not drop by more than 1\% for any of the datasets under any short-circuiting operation. That is, JustAsk model does not learn to align and fuse the modalities across the datasets. 

While the relative performance drop in the \emph{classical} language-only and video-only settings for JustAsk and FrozenBiLM on ActivityNet-QA, MSRVTT-QA and ATP-H is similar, QUAG, with combined dataset-model analysis, points out the differences in their  representations.

 The diagnostic results from QUAG provide confidence in achieving efficiency gains without significant performance drops by using QUAG-attention.  We find that applying QUAG-attention on FrozenBiLM and JustAsk reduces the number of multiplication operations by \textbf{13.6\%} and \textbf{68.0\%} respectively, for a less than 1\% drop in performance consistently for all the datasets. This raises serious concerns because this would mean that the models can learn to \emph{hack} their way around the accuracy metrics by leveraging shortcuts. Therefore, high accuracy on current datasets is not representative of multimodal understanding because the datasets contain exploitable biases that weaken the modality coupling.
\section{Can VideoQA Models Learn Highly-Coupled Multimodal Representations?}
\label{clavi}
In the previous sections we found that the models rely on different features on different datasets (Section \ref{exp_set}). However, unlike the simulated datasets in Section \ref{sec:simulation_study},  we do not know the degree of coupling between and within the  modalities for real world datasets. Then, \textbf{how can we test the performance of current VideoQA models in tightly-coupled multimodal settings?} The results from QUAG and QUAG-attention suggest that the current VideoQA datasets might have many biases and hence might not be a good test of multimodal understanding. Therefore, we build CLAVI, a simple VideoQA {stress-test} dataset by augmenting real world videos and their associated questions to get their respective temporal complements, and exhaustively cover all the possible combinations, ensuring high multimodal coupling and coverage. CLAVI is not positioned to replace existing datasets but rather to supplement them for enhancing the understanding of VideoQA models.

\begin{figure*}[h!]
  \centering
  \includegraphics[scale=0.57]{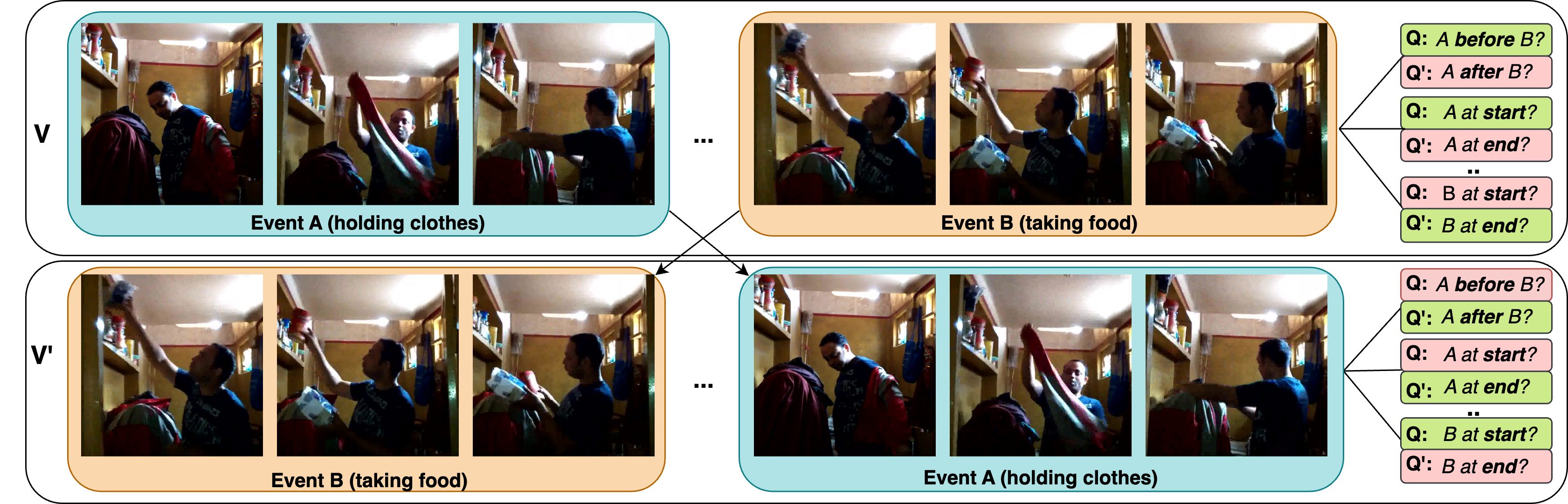}
  \caption{Illustrative example of the creation of CLAVI. In the original video (V), the action ``holding clothes'' (Event A; blue pane) follows ``taking food'' (Event B; brown pane). To create a complement video (V'), we swap the action segments without manipulating the  segment separating them. The questions (Q), along with their complement (Q'), are curated for each of the videos. Note that the color of the question panel reflects the correct answer (green for ``yes'', pink for ``no''). We provide the list of questions in Table \ref{table:clavi_eg}.}
  \label{fig:clavi}
\end{figure*}
\subsection{CLAVI: Testing Through Complements}
CLAVI consists of 6,018 videos and 114,342 questions (72,770 train and 41,572 test). It contains simple yes-no questions to probe the absolute temporal location of a single action (beginning/end) or the occurrence sequence for a pair of non-overlapping actions (before/after). CLAVI  allows for systematic testing of coupled multimodal understanding through balanced video and question temporal complements. We use question templates to automatically curate the question-answer pairs from the temporal grounding annotations of Charades-STA \citep{gao2017tall}. To create temporal complements  in the question domain, we replace  \emph{before} with \emph{after} and \emph{beginning} with \emph{end} and vice versa. Further, we create temporal video complements by swapping only the action-segments in the video (Figure \ref{fig:clavi}).

Many stress-test and diagnostic datasets have artefacts or discontinuous features \citep{wang2023paxion, zhang2020putting, bomatter2021pigs}, like the two points of discontinuity in the complement video of CLAVI. However, we ensure that it does not serve as a bias by maximizing the coverage of the dataset by exhaustively considering all the compositions of complements in video and question domains. This is also verified in our results in the following section where we test models with low, medium and high frame sampling rates on CLAVI. The exhaustivity also ensures tighter coupling within and between the modalities, as explained below.

\begin{table*}
    \centering
    \caption{List of questions and their complements in CLAVI for the illustrative example in Fig. \ref{fig:clavi} (E: Existence, BE:Beginning/End, BA:Before/After, and NC:Negative Control). NC questions contain an action (``washing mirror'', here) that never occurs in the video and hence, the answer is always ``no''. For brevity, we present 4 (out of 8) NC questions for BA type; comprehensive list in \S \ref{sec:list_questions}.}
    \begin{tabular}{l | l | l}
    \toprule
        Type & Question (Q) & Complement Question (Q') \\ 
        \midrule
        \multirow{2}{*}{E} & Was someone holding clothes? & ~  \\ 
        ~ & Was someone taking food? & ~  \\ \hline
        E-NC & Was someone washing mirror? & ~  \\ \hline
        \multirow{2}{*}{BE} & Was the person holding clothes at the \textbf{beginning}? &  Was the person holding clothes at the \textbf{end}? \\ 
        ~ & Was the person taking food at the \textbf{end}?  & Was the person taking food at the \textbf{beginning}? \\ \hline
        \multirow{2}{*}{BA} & Did holding clothes happen \textbf{before} taking food? & Did holding clothes happen \textbf{after} taking food? \\
        ~ & Did taking food happen \textbf{after} holding clothes? & Did taking food happen \textbf{before} holding clothes? \\ \hline
        \multirow{2}{*}{BA-NC} & Did holding clothes happen \textbf{before} washing mirror? & Did holding clothes happen \textbf{after} washing mirror? \\ 
        ~ & Did taking food happen \textbf{after} washing mirror? & Did taking food happen \textbf{before} washing mirror? \\
        \bottomrule
    \end{tabular}
    \label{table:clavi_eg}
\end{table*}

We briefly explain the design principle of CLAVI. We choose temporal sequence complements to test coupled multimodal understanding because it requires \textbf{unimodal understanding} within the modalities (sensitive to the sequence of \textit{(i)} frames in the video; \textit{(ii)} objects, verbs and temporal phrases in the question) as well as  \textbf{crossmodal understanding} (relating the sequence of actions in the video with that of the question). This also makes temporal ordering as one of the \textbf{fundamental} elements of VideoQA. Using yes-no questions with balanced negative instances allows us to have questions that are \textbf{unambiguous}, and answers that are \textbf{mutually exclusive} and \textbf{equally informative} to not be eliminated by prior biased knowledge. We deliberately maintain a simple design for question templates and answer vocabulary that excludes other abilities such as language comprehension, commonsense reasoning, and long-term memory to facilitate \textbf{isolated analysis and testing} of joint multimodal understanding. Also, we ensure that the dataset size is sufficiently large, as compared to the existing datasets, so that the models do not overfit (\S \ref{sec: clavi_dataset_comp}). 

Based on the temporal cue in the question, CLAVI contains three question types -- \textbf{Existence (E)}, \textbf{Beginning/End (BE)} and \textbf{Before/After (BA)}. Further, we define negative control questions containing actions that do not occur in the video (that is, the answer is always ``no'') for E and BA types as shown in Table \ref{table:clavi_eg}. Answering the negative control does not require understanding temporal cues. Hence, it serves the dual purpose of sanity check of learning and a baseline for learning by temporal shortcuts (low coupling instances).  We remove the bias against \emph{beginning} and \emph{end} by randomly extending the boundaries of the action-segments in the video (detailed curation process in \S \ref{sec:clavi_creation}). 


We want to evaluate the sensitivity of the model to the temporal cues in language and video separately. Hence, we define consistent \textbf{video-consistent accuracy ($\textrm{CAcc}_\mathcal{V}$)} and \textbf{text-consistent accuracy ($\textrm{CAcc}_\mathcal{T}$)}. 
\begin{gather*}
\textrm{CAcc}_\mathcal{V} = \mathds{1}[(y_{(V,Q)} == {\hat{y}}_{(V,Q)}) \wedge (y_{(V',Q)} == {\hat{y}}_{(V',Q)})]  \\
\textrm{CAcc}_\mathcal{T} = \mathds{1}[(y_{(V,Q)} == {\hat{y}}_{(V,Q)}) \wedge (y_{(V,Q')} == {\hat{y}}_{(V,Q')})] 
\end{gather*} 
 where $y$ and $\hat{y}$ are the ground truth and predictions, with the video and question indicated in the subscript and $\mathds{1}$ is the indicator function. Note that ($'$) indicates the complement in CLAVI. We average the consistent accuracies  over the entire dataset and report it separately for the \textbf{control subset} (E, E-NC, and BA-NC question types) and the \textbf{complement subset} (BE and BA question types). The coupling between the modalities is lower in the control subset since can be answered by even ignoring the temporality in video and text while answering the complement subset (higher modality coupling) necessitates joint multimodal understanding.

\subsection{Experiment and Analysis}
\begin{table*}[htb!]
  \caption{Test performance in terms of our consistent accuracy metric (CAcc) on CLAVI after finetuning. The control subset can be answered by ignoring temporality, and hence has low multimodal coupling while the complement subset has high multimodal coupling (red color denotes near-trivial performance on complement subset and green denotes high performance (the more the better)).}
  \label{clavi-ft-acc}
  \centering
  \begin{tabular}{l c c c c}
    \toprule
    Metric & JustAsk & FrozenBiLM & Singularity-T & All-In-One+  \\
    ~ &  \footnotesize{\citep{yang2021justask}} & \footnotesize{\citep{yang2022zero}} & \footnotesize{\citep{lei2022revealing}} & \footnotesize{\citep{wang2023all}} \\
    \midrule
    Balanced Acc & 72.2~\textpm~0.2 & 80.5~\textpm~0.1 & 76.8~\textpm~0.5 & 73.9~\textpm~0.1 \\ \hline
    $\textrm{CAcc}_{\mathcal{V}}$ & 50.6~\textpm~0.3 & 74.0~\textpm~0.1 & 47.2~\textpm~1.1 & 49.6~\textpm~0.5 \\ 
    $\textrm{CAcc}_{\mathcal{T}}$ & 50.3~\textpm~0.1 & 75.5~\textpm~0.1 & 47.0~\textpm~1.0 & 49.5~\textpm~0.3 \\ \hline
    $\textrm{CAcc}_{\mathcal{V}}$-control & 98.0~\textpm~0.2 & 93.2~\textpm~0.2 & 92.7~\textpm~2.0 & 98.1~\textpm~0.5 \\ 
    $\textrm{CAcc}_{\mathcal{T}}$-control & 98.2~\textpm~0.2 & 93.7~\textpm~0.2 & 93.5~\textpm~1.9 & 98.2~\textpm~0.7 \\ \hline
    $\textrm{CAcc}_{\mathcal{V}}$-complement & \textcolor{red}{\phantom{0}\textbf{3.6~\textpm~0.1}} & \textcolor{ForestGreen}{\textbf{54.1~\textpm~0.2}} & \textcolor{red}{\phantom{0}\textbf{1.7~\textpm~0.2}} & \textcolor{red}{\phantom{0}\textbf{1.2~\textpm~0.3}} \\ 
    $\textrm{CAcc}_{\mathcal{T}}$-complement & \textcolor{red}{\phantom{0}\textbf{2.4~\textpm~0.1}} & \textcolor{ForestGreen}{\textbf{57.2~\textpm~0.2}} & \textcolor{red}{\phantom{0}\textbf{0.5~\textpm~0.2}} & \textcolor{red}{\phantom{0}\textbf{0.8~\textpm~0.1}} \\
    \bottomrule
  \end{tabular}
\end{table*}

We finetune and evaluate 4 models: JustAsk~\citep{yang2021justask} (640 frames), FrozenBiLM~\citep{yang2022zero} (10 frames), Singularity-Temporal~\citep{lei2022revealing} (12 frames) and All-In-One+~\citep{wang2023all} (3 frames) on CLAVI using the official finetuning instructions (\S \ref{sec:clavi_exp_details_supp}). We do not report zero-shot results because the models almost always answer either of the two answers, which is a well known phenomenon in multimodal QA \citep{guo2021loss}. We follow the same experimental settings as discussed in Section \ref{exp_set}. To account for class imbalance in the answers, we use balanced accuracy for validation and testing.

The results are summarized in Table \ref{clavi-ft-acc}. All the models have greater than 70\% balanced accuracy. Next, we do a finegrained analysis of the performance.

Text and video consistent accuracies are greater than 90\% 
for
the control 
subset
for all the models. This is because, unlike the complement subset, the control subset requires less coupled understanding. That is, the model can answer it correctly by simple shortcuts -- irrespective of the context of the negative control action in the question and the location of the object and/or the action in the video. However, for achieving high consistent accuracies on the complement subset, the model needs to jointly understand the order of the events and the temporal cues in the question along with the order of the events in the video. We get significantly lower consistent accuracies (less than 4\%) for the complement subset, except for FrozenBiLM. Overall, \textbf{most models have trivial performance on high-multimodal coupling subset}. 

No dataset can be free of biases. Therefore, while low performance on CLAVI is representative of poor coupled multimodal understanding, high performance doesn't necessarily guarantee it. How can we be sure that FrozenBiLM is not learning spurious shortcuts on CLAVI? We find that the video-average QUAG-attention on FrozenBiLM cause the $\textrm{CAcc}_{\mathcal{T}}$-complement and $\textrm{CAcc}_{\mathcal{V}}$-complement to drop to \textbf{23\%} and \textbf{3.6\%} respectively. That is, the performance on the complement subset significantly drops under multimodal impairment. However, $\textrm{CAcc}_{\mathcal{T}}$-control and $\textrm{CAcc}_{\mathcal{V}}$-control values increase to \textbf{98.6\%} and \textbf{99.2\%} respectively, perhaps because QUAG-attention promotes reliance on shortcuts, and the control subset can be solved easily by shortcuts.  These results confirm FrozenBiLM's high reliance on multimodal representations for its higher performance.

Beyond the consistency accuracy metrics we can use CLAVI for diverse representation analyses. As an example, we present  a qualitative representation sensitivity analysis for FrozenBiLM in \S \ref{sec:clavi_additional_analysis}. We align the attention matrices for complement pairs and find that the representations of correctly answered complement pairs are more distinct than the wrongly answered pairs to support our understanding.


\section{Related Work}
\label{sec:rel_work}
\textbf{Dataset Biases}: Works in NLP \citep{papadimitriou2022classifying, sinha2021masked}, vision \citep{brendel2018bagnets} and vision-language \citep{yuksekgonul2023when} demonstrate that models can achieve high performance without even understanding the sequence of the embeddings. This is partly because the current benchmarks have unintended biases that could potentially be exploited by models to learn shortcuts;  hence accuracy is not always a faithful metric \citep{pham2021out, yuksekgonul2023when, kafle2017analysis, sevilla2021only}. 
For VideoQA, MovieQA \citep{tapaswi2016movieqa} and TVQA \citep{lei2018tvqa} datasets are biased towards plot understanding or dialogue comprehension \citep{mbintvqa}. Biases are not always immediately apparent; for example, Social-IQ \citep{zadeh2019social} contains sentiment-biased annotations \citep{gat2021perceptual}. Moreover, statistical regularities like answer length, answer frequency \citep{Goyal_2017_CVPR, agrawal2016analyzing} and co-occurrence \citep{dancette2021beyond, manjunatha2019explicit, subramanian2019analyzing} introduce spurious features. Overall, these biases allow the models learn shortcuts \citep{geirhos2020shortcut} that circumvent multimodal reasoning \citep{chao2018being, ye2021case}. While synthetic VideoQA benchmarks such as VQuAD \citep{gupta2022vquad}, CLEVRER \citep{yi2019clevrer} have been carefully curated to mitigate many biases, they are unable to capture the intricate dynamics of the real world. Recently proposed Preception Test \citep{patraucean2023perception}, while comprehensive, does not contain diagnostic metrics that penalize the effect of shortcut temporal learning. 
We curate CLAVI by systematically augmenting real-world videos to faithfully represent the complexity of the physical world while controlling the biases to confidently evaluate multimodal temporal understanding.

\textbf{Shortcut Learning}: Tangential to the bias amelioration methods \citep{cadene2019rubi, clark-etal-2019-dont}, \citet{lei2022revealing} and \citet{mbintvqa} achieve state-of-the-art performance with simple models by leveraging VideoQA dataset shortcuts in the model. ATP \citep{buch2022revisiting} demonstrates single frame bias  by re-training the models with an informative frame-selection module to achieve competitive performance. Perceptual Score \citep{gat2021perceptual} quantifies modality bias in terms of relative performance drop under modality-permutation operation. 
QUAG combines these ideas to evaluate the dependence of models on shortcuts for circumventing multimodal understanding in terms of performance drop under multimodal representation collapse. 
Unlike others, it assists in identifying sub-optimal representations in a combined model-dataset approach at test time.

\textbf{Leveraging Complements}: We share our motivation for developing CLAVI with VQA-CP \citep{agrawal2018don}: that iid train-test splits in the presence of strong priors leads to learning via shortcuts. However, rather than reducing the bias by mining new complementary image instances, CLAVI weakens prior of multimodal understanding with synthesized balanced video-question temporal hard-negatives. 
\citet{momeni2023verbs} and \citet{wang2023paxion} have employed hard-negatives for improving verb-understanding in VideoQA models.
CLAVI is inspired by \citet{bagad2023testoftime} stitch video clips to improve the temporal understanding of video-language models. Unlike previous works, we (i) introduce atemporal question types, and (ii) propose a new diagnostic metric (consistent accuracy) that are needed for understanding multimodality in VideoQA models. 

{\textbf{Multimodal Fusion Interpretability and Visualization: } \citet{liang2022foundations} and \citet{liang2023multimodal} analyze multimodal fusion interactions along the dimensions of response, information, and mechanics. The closest alignment of QUAG and CLAVI is at the interface of multimodal fusion response and mechanics. Previous works have quantified the presence or absence of specific kinds of modality interactions through the study of datasets \citep{dancette2021beyond}, models \citep{chefer2021generic}, projections onto simpler models \citep{hessel2020does, wortwein2022beyond} and visualization studies \citep{liang2022multiviz, aflalo2022vl, wang2021m2lens}. However, using QUAG, unlike other methods, we perform a combined dataset-model analysis without any additional parameters or finetuning.}

\section{Conclusion}
\label{conclusion}
In this work, we perform a rigorous analysis of VideoQA models, focusing on multimodal representations. 
We introduced QUAG, a coupled dataset-model approach, to conduct a systematic analysis of learned multimodal representations. It provides deep insights into \emph{how} the models infer and \emph{why} the models fail. We found that VideoQA models do not necessarily learn  to align and fuse the information both -- within and between the modalities from multimodal datasets.  Using this understanding, we developed QUAG-attention and exposed the sub-optimality of VideoQA models. Since the datasets might not have high coupling, we proposed CLAVI, a stress-test dataset for testing highly-coupled multimodal understanding in VideoQA models. With the simple task of temporal ordering we find that most of the current models are unable to jointly infer from text and video modalities.

All our proposed approaches -- QUAG, QUAG-attention and CLAVI are simple, compute-friendly and generic to be extended to any combination of modalities, datasets and models. Our thorough and systematic dataset-model combined representation analysis asserts the inability of VideoQA models to learn highly-coupled multimodal settings, that is not evaluated by current datasets.


\section*{Acknowledgements}
This work was supported by an A*STAR CRF award to C.T. and the National Research Foundation, Singapore, under its NRF Fellowship (NRF-NRFF14-2022-0001) to B.F.
\section*{Impact Statement}
Datasets in machine learning have a history of containing unintentional biases  like race, gender, age along with safety and privacy concerns \citep{birhane2021large, peng2021mitigating, hirota2022gender}. We curate CLAVI from existing and popular Charades \citep{sigurdsson2016hollywood} dataset because it is well-studied, licensed for academic usage, and collected in controlled settings with consent. Further the owners of Charades ensure the anonymity and privacy of the participants. Also, for automatically generated questions, we ensure to keep the question templates gender neutral. One of the central goals of our work was to reassert the brittleness in multimodal models by presenting a combined dataset-model centric interpretable representation learning approach through QUAG and CLAVI. We hope our work galvanizes the research community further to not just blindly trust the accuracy score on benchmarks but thoroughly investigate the potential biases that are \emph{(1)} present in the dataset and \emph{(2)} are learned by the models. 
\bibliography{bibliography}
\bibliographystyle{icml2024}

\newpage
\appendix
\onecolumn
\section{QUAG}
\subsection{Toy Example}
\label{sec:toy_example_sc}
\begin{figure}[h!]
  \centering
  \includegraphics[width=\textwidth]{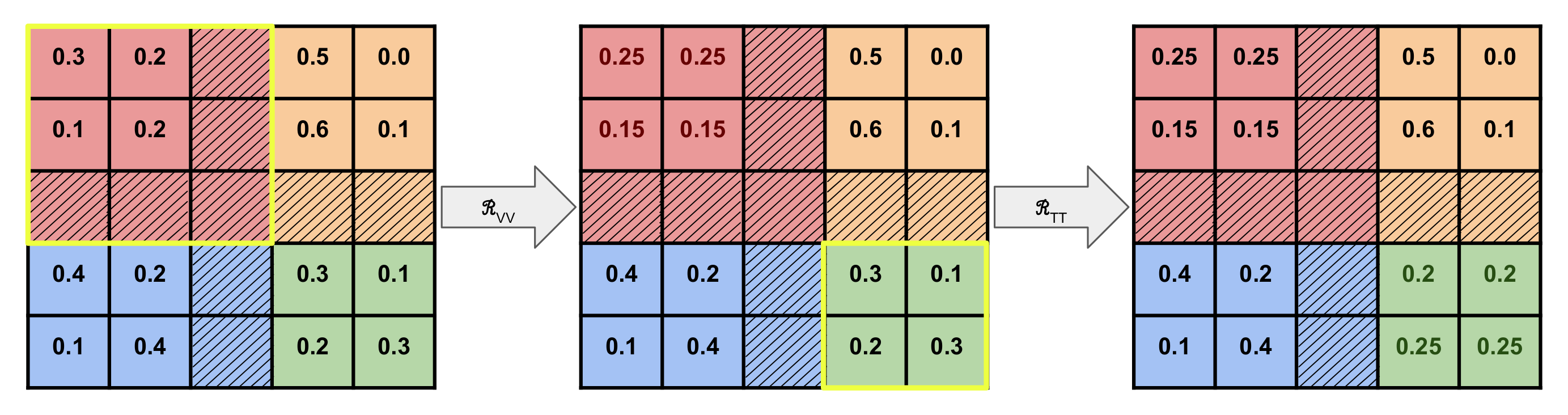}
  \caption{Toy example of $\phi(Z, [\mathcal{TT, VV}])$, where $Z$ is the input (left-most matrix) $\mathcal{R}$ is the row-wise average and replace operator and hatching denotes padding. The quadrants that are operated on are highlighted in bright yellow box. Note that $L_\mathcal{V} = 3$ and $L_\mathcal{T} = 2$, such that video embeddings are pre-concatenated to question embeddings (as in the main manuscript). The cells are colored as per their quadrants ($\mathcal{VV: \textrm{red}}, \mathcal{VT: \textrm{yellow}}, \mathcal{TV: \textrm{blue}}, \mathcal{TT: \textrm{green}}$)}
  \label{fig:quag_example}
\end{figure}

Consider the toy example in Fig \ref{fig:quag_example}. The left-most matrix is the input matrix. As per the definition of $\phi$, we can write, $\phi(Z, [\mathcal{TT, VV}]) = \mathcal{R_{TT}} \circ \mathcal{R_{VV}}(Z)$ ({which we define as unimodal short-circuiting operation}). We demonstrate the successive application of $\mathcal{R}$ operator in the example. Note that the padding is ignored; this is equivalent to applying $\mathcal{R}$ to the padding-free sub-partition of the quadrant. Also, as illustrated in the example, since the quadrants cannot overlap, the sequence of application of $\mathcal{R}$ does not matter. 
\label{sec:lemma_proof}
\subsection{Proof of \cref{lem:rank_reduction}}
\textbf{\cref{lem:rank_reduction}.} \emph{For an attention matrix $\bm A$ of size $(l_\mathcal{V} + l_\mathcal{T}) \times (l_\mathcal{V} + l_\mathcal{T})$, the rank, $\rho$, after video short-circuiting is bounded as: $\rho(\phi(\bm A, [\mathcal{VV, TV}])) \leq l_\mathcal{T} + 2$.}

\begin{proof}
    The attention matrix $\bm A$ can be written in terms of its partitions as:
     $\begin{pmatrix}
  A_{\mathcal{VV}} & A_{\mathcal{VT}}\\ 
  A_{\mathcal{TV}} & A_{\mathcal{TT}}
\end{pmatrix}$. Also, $\rho(\bm A) \leq l_\mathcal{V} + l_\mathcal{T}$
Then, $\phi(\bm A, [\mathcal{VV, TV}]) = \begin{pmatrix}
  \begin{pmatrix}
  \mu_{1\phantom{+l_\mathcal{V}}}  & \cdots & \mu_{1\phantom{+l_\mathcal{V}}}\\ 
  \cdot & \cdots & \cdot \\
  \cdot & \cdots & \cdot \\
  \mu_{l_\mathcal{V}\phantom{+l_\mathcal{V}}} & \cdots  & \mu_{l_\mathcal{V}\phantom{+l_\mathcal{V}}}
\end{pmatrix} & A_{\mathcal{VT}}\\ \\
  \begin{pmatrix}
  \mu_{l_{\mathcal{V}}+1} & \cdots  & \mu_{l_{\mathcal{V}}+1}\\ 
  \cdot & \cdots & \cdot \\
  \cdot & \cdots & \cdot \\
  \mu_{l_{\mathcal{T}}+l_\mathcal{V}} & \cdots  & \mu_{l_{\mathcal{T}}+l_\mathcal{V}}
\end{pmatrix} & A_{\mathcal{TT}}
\end{pmatrix}$, where $\mu_i$ is the mean of the $i^{\textrm{th}}$ row of the partition.

Assuming $\mu_1 \neq 0$, we apply the following row operation on the matrix: 
$R_j \leftarrow R_j - \frac{\mu_j}{\mu_1}R_1$, for $j = 2, 3, \cdots, (l_\mathcal{V}+l_\mathcal{T})$, where $R_k$ represents $k^{\textrm{th}}$ row of $\phi(\bm A, [\mathcal{VV, TV}])$. The resulting matrix is:
$\begin{pmatrix}
  \begin{pmatrix}
  \mu_{1}  & \cdots & \mu_{1}\\ 
  \cdot & \cdots & \cdot \\
  0 & \cdots & 0 \\
  0 & \cdots  & 0
\end{pmatrix} & \tilde{A}_{\mathcal{VT}}\\ \\
  \begin{pmatrix}
  0_{\phantom{1}} & \cdots  & 0_{\phantom{1}} \\ 
  \cdot_{\phantom{1}} & \cdots & \cdot_{\phantom{1}} \\
  0_{\phantom{1}} & \cdots & 0_{\phantom{1}} \\
  0_{\phantom{1}} & \cdots  & 0_{\phantom{1}}
\end{pmatrix} & \tilde{A}_{\mathcal{TT}}
\end{pmatrix} 
$

Rearranging the partitions, we can write,

\begin{equation*}
\begin{pmatrix}
  \begin{pmatrix}
  \mu_{1}  & \cdots & \mu_{1}\\ 
\end{pmatrix} & \hat{A}_{\mathcal{VT}}\\ \\
  \begin{pmatrix}
  0_{\phantom{1}} & \cdots  & 0_{\phantom{1}} \\ 
  \cdot_{\phantom{1}} & \cdots & \cdot_{\phantom{1}} \\
  0_{\phantom{1}} & \cdots & 0_{\phantom{1}} \\
  0_{\phantom{1}} & \cdots  & 0_{\phantom{1}}
\end{pmatrix} & \hat{A}_{\mathcal{TT}}
\end{pmatrix} \equiv 
\begin{pmatrix}
   [\hat A_{\mathcal{VV}}]_{1 \times l_\mathcal{V}} & [\hat  A_{\mathcal{VT}}]_{1 \times l_\mathcal{T}}\\ 
  [\hat  A_{\mathcal{TV}}]_{(l_\mathcal{V}+l_\mathcal{T}-1) \times l_\mathcal{V}} & [\hat  A_{\mathcal{TT}}]_{(l_\mathcal{V}+l_\mathcal{T}-1) \times l_\mathcal{T}}
\end{pmatrix} \equiv
\hat{\bm A}
\end{equation*} where $[.]_{a \times b}$ denotes that the shape of the partition is $a \times b$.

Using the rank-inequality result of \citet{meyer1973generalized}, we can write:
\begin{equation}
\label{eq:rank_ineq}
\rho(\bm{\hat{ A}}) \leq \rho(\hat A_{\mathcal{VV}}) + \rho(\hat  A_{\mathcal{TV}}) + \rho(\hat  A_{\mathcal{VT}}) + \rho(\hat  A_{\mathcal{TT}} - \hat  A_{\mathcal{TV}} \hat A_{\mathcal{VV}}^{-} \hat  A_{\mathcal{VT}})
\end{equation} 
where $\hat A_{\mathcal{VV}}^{-}$ denotes the Moore-Penrose pseudo-inverse of the partition. Since $\hat A_{\mathcal{TV}}$ is a zero matrix, $\rho(\hat A_{\mathcal{TV}}) = 0$, and $\rho(\hat  A_{\mathcal{TT}} - \hat  A_{\mathcal{TV}} \hat A_{\mathcal{VV}}^{-} \hat  A_{\mathcal{VT}}) = \rho(\hat  A_{\mathcal{TT}}) \leq l_{\mathcal{T}}$ (we assume that $l_\mathcal{V} > 1$ for all practical purposes). Also, $\rho(\hat A_{\mathcal{VV}}) = 1$, and $\rho(\hat A_{\mathcal{VT}}) \leq 1$. Plugging these values in \cref{eq:rank_ineq}, we get, $\rho(\bm{\hat{ A}}) \leq 1 + 0 + 1 + l_\mathcal{T} \implies \rho(\bm{\hat {A}}) \leq l_\mathcal{T} + 2$. 

Since row operations do not change column rank, $\rho(\bm {\hat {A}}) = \rho(\phi(\bm A, [\mathcal{VV, TV}])) \textrm{. Therefore, } \rho(\phi(\bm A, [\mathcal{VV, TV}])) \leq l_\mathcal{T} + 2$

\end{proof}

\subsection{Attention Map Visualization}
\begin{figure}[ht!]
  \centering
  \includegraphics[width=\textwidth]{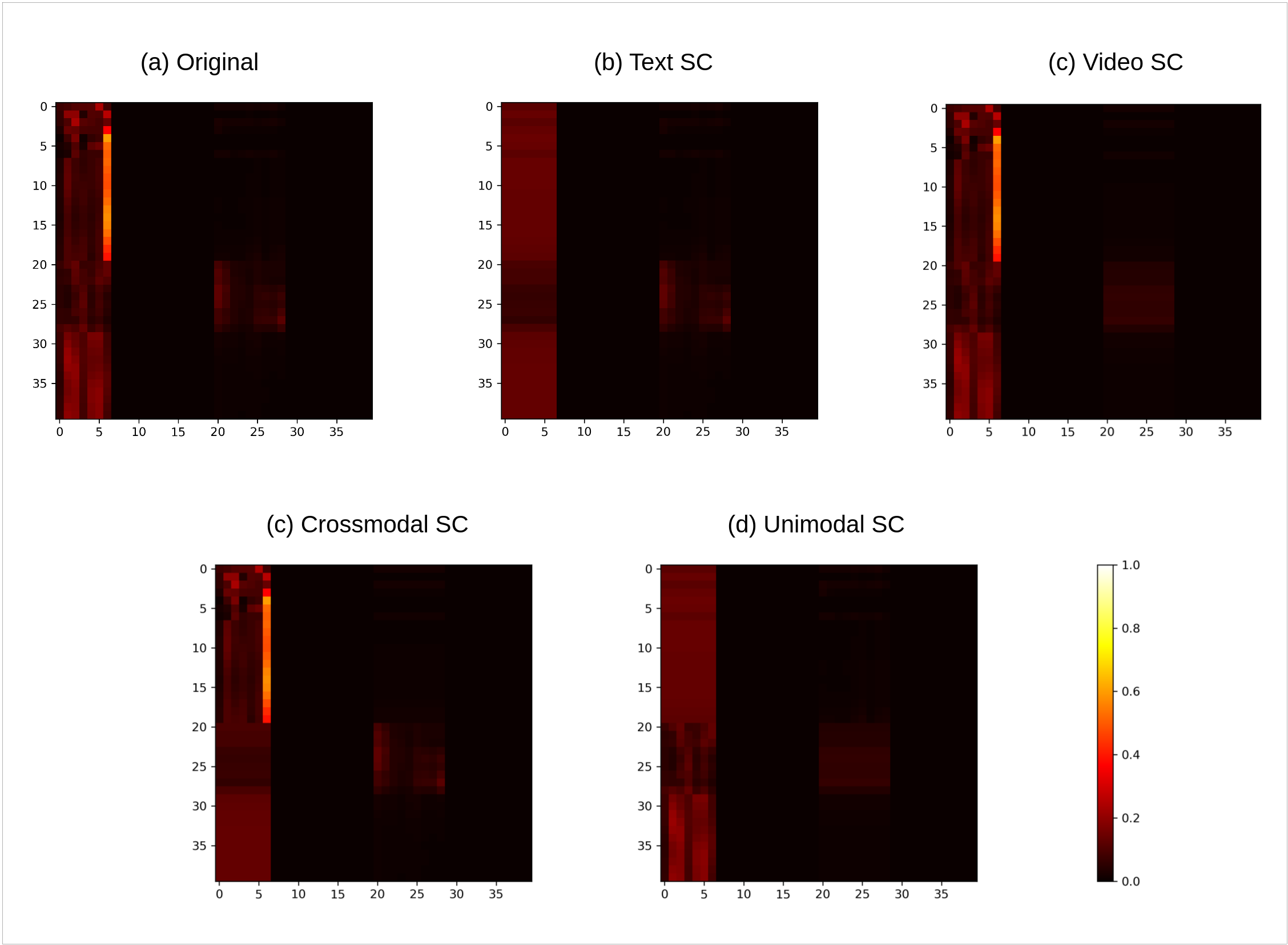}
  \caption{Visualization of the first attention head, as a heatmap, from the second layer of JustAsk model with $l_\mathcal{V} = 20$ and $l_\mathcal{T} = 20$. Note that here the text embeddings are pre-concatenated to the video embedding in the input. The lengths of the video and text tokens are 9 and 7 respectively. The text and video tokens are individually padded to length 20 each. We visualize \textit{(a)} the original attention values and \textit{(b)-(d)} after short-circuiting (SC) operations. }
  \label{fig:quag_sc_visual}
\end{figure}

We provide a visualization example of the attention values before and after short-circuting operations in Figure \ref{fig:quag_sc_visual}.
\subsection{Code}
\label{sec:quag_pseudo_code}
Below is the implementation of QUAG as an augmentation of the existing self-attention function. We use row-wise average and replace operation in each if-clause statements, while ignoring the padding, to ablate the effect of the quadrant. 
\begin{lstlisting}[language=Python]
def self_attention(inputs, mask, dim_model, l_v, l_t, quads):
    # Inputs:
    #   inputs: Tensor of shape (batch_size, sequence_length, dim_model)
    #   mask: Tensor of shape (batch_`size, sequence_length)
    #   dim_model: Dimension of the model (e.g., 512)
    #   l_v: int    maximum length of video tokens
    #   l_t: int    maximum length of question tokens
    #   quads: list containing elements from {'VV', 'VT', 'TV', 'TT'}
    query = linear_transform_query(inputs)
    key = linear_transform_key(inputs)
    value = linear_transform_value(inputs)
    attention_scores = compute_attention_scores(query, key, mask)
    apply_quag(attention_scores, mask, l_v, l_t, quads)
    attended_output = apply_attention_scores(attention_scores, value)
    return attended_output

def compute_attention_scores(query, key, mask):
    scaled_dot_product = dot_product(query, key) / sqrt(dim_model)
    attention_scores = softmax(scaled_dot_product + (1 - mask) * -1e9)
    return attention_scores

def apply_quag(attention_scores, mask, l_v, l_t, quads):
    if 'VV' is in quads:
        replace_with_rowwise_average(attention_scores[:, :l_v, :l_v], mask[:, :l_v, :l_v])
    if 'VT' is in quads:
        replace_with_rowwise_average(attention_scores[:, :l_v, -l_t:], mask[:, :l_v, -l_t:])
    if 'TV' is in quads:
        replace_with_rowwise_average(attention_scores[:, -l_t:, :l_v], mask[:, -l_t:, :l_v])
    if 'TT' is in quads:
        replace_with_rowwise_average(attention_scores[:, -l_t:, -l_t:], mask[:, -l_t:, -l_t:])

def replace_with_rowwise_average(scores, mask):
    rowwise_sum = sum(scores, axis=-1)
    rowwise_mean = rowwise_sum / sum(mask, axis=-2)
    expanded_rowwise_mean = expand_dims(rowwise_mean, axis=-1)
    replace_elements(scores, expanded_rowwise_mean)

def apply_attention_scores(attention_scores, value):
    attended_output = dot_product(attention_scores, value)
    return attended_output
\end{lstlisting}

Next, we provide the code for QUAG-attention. QUAG-attention modifies the existing self-attention block in the fusion module by replacing the block with the block average. We also demonstrate the normalizing the softmax function so that the each single average sequence is representative of the constituent  sequences.

\begin{lstlisting}[language=Python]
def quag_attention(inputs, mask, dim_model, l_v, l_t, type):
    # Inputs:
    #   inputs: Tensor of shape (batch_size, sequence_length, dim_model)
    #   mask: Tensor of shape (batch_size, sequence_length)
    #   dim_model: Dimension of the model (e.g., 512)
    #   l_v: int    maximum length of video tokens
    #   l_t: int    maximum length of question tokens
    #   type: one of 'text', 'video', 'text-video'
    query = linear_transform_query(inputs)
    avg_input = compute_avg_input(inputs, l_v, l_t, type)
    key = linear_transform_key(avg_input)
    value = linear_transform_value(avg_input)
    mask = apply_mask(mask, l_v, l_t, type)
    scaled_dot_product = compute_scaled_dot_product(query, key, dim_model, mask)
    attention_scores = softmax(scaled_dot_product)
    attended_output = apply_attention_scores(attention_scores, value)
    return attended_output

def compute_avg_input(inputs, l_v, l_t, type):
    if type == "video":
        avg_upper_block = sum(inputs[:, :l_v, :], axis=-2)
        avg_upper_block = expand_dims(avg_upper_block, axis=1)
        avg_input = concatenate((avg_upper_block, inputs[:, :-l_t, :]), axis=1)
    elif type == "text":
        avg_lower_block = sum(inputs[:, :-l_t, :], axis=-2)
        avg_lower_block = expand_dims(avg_lower_block, axis=1)
        avg_input = concatenate((inputs[:, :l_v, :], avg_lower_block), axis=1)
    elif type == "text-video":
        avg_upper_block = sum(inputs[:, :l_v, :], axis=-2)
        avg_upper_block = expand_dims(avg_upper_block, axis=1)
        avg_lower_block = sum(inputs[:, :-l_t, :], axis=-2)
        avg_lower_block = expand_dims(avg_lower_block, axis=1)
        avg_input = concatenate((avg_upper_block, avg_lower_block), axis=1)
    return avg_input

def apply_mask(mask, l_v, l_t, type):
    mask = expand_dims(mask, axis=-1)
    mask = tile(mask, [1, 1, sequence_length])
    
    if "video" in type:
        video_length = sum(mask[:, :l_v, 0], axis=1)
        video_length = expand_dims(video_length, axis=-1)
        scaled_dot_product[:, :, 0] = scaled_dot_product[:, :, 0] * log(video_length)
        upper_mask = ones(mask.shape[0], mask.shape[1], 1)
        mask = concatenate((upper_mask, mask[:, :, l_v:]), axis=-1)

    if "text" in type:
        text_length = sum(mask[:, :-l_t, 0], axis=1)
        text_length = expand_dims(text_length, axis=-1)
        scaled_dot_product[:, :, -1] = scaled_dot_product[:, :, -1] * log(text_length)
        lower_mask = ones(mask.shape[0], mask.shape[1], 1)
        mask = concatenate((mask[:, :, :-l_t], lower_mask), axis=-1)
    
    return mask

def compute_scaled_dot_product(query, key, dim_model, mask):
    scaled_dot_product = dot_product(query, key) / sqrt(dim_model)
    return scaled_dot_product

def apply_attention_scores(attention_scores, value):
    attended_output = dot_product(attention_scores, value)
    return attended_output

\end{lstlisting}
\subsection{Training and Data Details for Simulation Study}
\label{sec:supp_sim}
Using the dataset generation strategy as described in the main paper, we sample 40,000 data points (24,000: training, 8,000: validation, 8,000: testing). We train a 4-layer transformer model, with  learnable modality encoding and sinusoidal position encoding, having a dimensionality of 100. Following conventions, we set the hidden dimension to four times the hidden dimension (400). To prevent overfitting, we add dropouts in the embedding, attention, and penultimate layers. We Adam optimizer with a learning rate of 0.001 for 2000 epochs. The training batch size was 1024. 
\subsection{Experiment Details for Real World Data}
\label{sec:quag_impl_detail}

All our experiments were performed on 4 NVIDIA A5000 GPUs. We use the official open-source code of the models on GitHub and modify only the self-attention modules. We use the official evaluation code and checkpoints.  For NeXT-QA, we use the official dataset and finetune the models with the default parameters 

As mentioned in the main manuscript, we use the official checkpoints and code of JustAsk \href{https://github.com/antoyang/just-ask}{[website]} and FrozenBiLM \href{https://github.com/antoyang/FrozenBiLM}{[website]}. For all the experiments with JustAsk, we use the checkpoints of the model pretrained on HowToVQA69M and WebVidVQA3M. For FrozenBiLM, we use the WebVid10M-pretrained checkpoint for all our experiments. Since QUAG operates at inference time, we do not need to perform any training. Since the model owners do not report results on NeXT-QA, we finetune the models with the official recipe to achieve performance similar to that independently reported by others \cite{xiao2022video}. While FrozenBiLM can also take subtitles as the input, for fair comparison, we do not pass it in any of the experiments. We provide the hardware details in the main manuscript.
\subsection{Finegrained Accuracies}
 \subsection{JustAsk Model}
We present the fine-grained performance of JustAsk on the discussed datasets in Tables \ref{justask-aqa}, \ref{justask-msrvtt},  \ref{justask-nextqa}, and \ref{justask-nextqa-atph}
\begin{table}
  \caption{Fine-grained performance of JustAsk on ActivityNet-QA}
  \label{justask-aqa}
  \centering
  \begin{tabular}{lrrrrrrrrr}
    \toprule
    Config & Motion &  Spatial & Temp & Y/N & Color & Obj & Loc & Num & Other  \\
    \midrule
    Baseline &  30.6  & 19.9  & 4.9  &  64.2 &  34.7 &  26.7 & 35.5  &  48.9 &  36.8    \\
    Lang-only    & 1.4  &  9.1 &  4.3 & 51.8  & 28.7  & 23.0  & 16.6  & 46.9  & 29.1  \\
    Vid-only    & 20.3  & 0.9  &  1.8 &  0.0 & 0.0  & 1.6  &  1.3 & 0.0  & 0.7  \\
    SC: unimodal    & 30.1  &  19.1 & 4.9  &  63.9 &  33.6 & 26.4  & 36.8  & 48.4  & 37.0  \\
    SC: crossmodal    & 28,0  & 18.9  & 4.8  & 64.7  & 34.7  & 25.8  &  35.5 &  48.5 & 36.4  \\
    SC: text    &  30.4 &  19.3 & 5.0  &  64.1 &  34.0 & 26.4  & 35.5  & 46.7  & 37.2  \\
    SC: video    &  28.6 & 18.8  & 4.5  & 64.3  &  34.6 &  25.5 &  35.5 & 48.4  & 36.1  \\
    QUAG-attention    &  28.1 &  18.5 & 4.9  &  64.1 &  33.6 & 25.2  &  34.7 &  48.0 & 36.6  \\
    \bottomrule
  \end{tabular}
\end{table}

\begin{table}[h!]
  \caption{Fine-grained performance of JustAsk on MSRVTT-QA}
  \label{justask-msrvtt}
  \centering
  \begin{tabular}{lrrrrrr}
    \toprule
    Config & What & How & Color & Where & Who & When   \\
    \midrule
    Baseline    &  35.8 &  83.7 &  51.7 &  39.4 &  51.3 & 82.3   \\
    Lang-only    &  24.3 & 83.3  & 43.4  &  30.5 & 37.1  & 72.3    \\
    Vid-only    &  8.5 & 0.0  &  3.5 & 0.4  &  3.0 & 10.1   \\
    SC: unimodal    & 35.6  & 83.3  &  51.8 &  39.8 & 50.8  & 82.3    \\
    SC: crossmodal    &  35.35 &  83.75 &  51.98 & 39.8  & 50.8  & 81.8     \\
    SC: text    &  35.7 & 83.2  & 51.8  &  39.0 & 50.8  & 82.1    \\
    SC: video    &  35.4 & 83.8  & 51.8  &  39.8 &  50.7 & 81.6   \\
    QUAG-attention    &  35.1 & 83.5  &  51.1 &  38.6 & 50.2  & 82.1    \\
    \bottomrule
  \end{tabular}
\end{table}
\begin{table}[h!]
  \caption{Fine-grained performance of JustAsk on NeXT-QA}
  \label{justask-nextqa}
  \centering
  \begin{tabular}{lrrr}
    \toprule
    Config & Causal & Temporal & Descriptive   \\
    \midrule
    Baseline    &  50.8 & 52.8  &  65.0     \\
    Lang-only    &  39.5 & 44.3  &  47.1     \\
    Vid-only    &  39.2 & 37.9  &  44.0    \\
    SC: unimodal    &  50.5 & 52.5  & 65.3     \\
    SC: crossmodal    &  50.8 & 51.8  & 65.0      \\
    SC: text    &  50.7 & 52.7 &  65.0      \\
    SC: video    &  50.7 & 52.1  & 65.0    \\
    QUAG-attention    & 50.8 & 52.0 & 65.1    \\
    \bottomrule
  \end{tabular}
\end{table}
\begin{table}[h!]
  \caption{Fine-grained performance of JustAsk on ATP-Hard subset of NeXT-QA}
  \label{justask-nextqa-atph}
  \centering
  \begin{tabular}{lrr}
    \toprule
    Config & Causal & Temporal   \\
    \midrule
    Baseline    &  44.4 & 43.4        \\
    Lang-only    & 41.2  &  43.1       \\
    Vid-only    & 23.5  &   22.3      \\
    SC: unimodal    & 43.2  &  43.3      \\
    SC: crossmodal    & 44.2  &  44.4       \\
    SC: text    &  43.7 &  43.4        \\
    SC: video    & 44.3  &  44.4     \\
    QUAG-attention    & 44.2  &  43.9     \\
    \bottomrule
  \end{tabular}
\end{table}

 \subsection{FrozenBiLM Model}
We present the fine-grained performance of FrozenBiLM on the discussed datasets in Tables \ref{frozenbilm-aqa}, \ref{frozenbilm-msrvtt},  \ref{frozenbilm-nextqa}, and \ref{frozenbilm-nextqa-atph}
\begin{table}
  \caption{Fine-grained performance of FrozenBiLM on ActivityNet-QA}
  \label{frozenbilm-aqa}
  \centering
  \begin{tabular}{lrrrrrrrrr}
    \toprule
    Config & Motion &  Spatial & Temp & Y/N & Color & Obj & Loc & Num & Other  \\
    \midrule
    Baseline    & 30.1  & 22.5  &  6.4 &  75.6 &  34.6 &  27.7 & 37.1  &  55.8 &  41.6 \\
    Lang-only    & 2.6  & 10.5  & 4.8  &  63.3 &  32.3 & 23.9  &  16.6 & 44.7  & 31.6  \\
    Vid-only    & 0.0  & 0.0  & 0.5  & 0.0  &  0.0 &  0.0 &  0.0 & 0.0  & 0.0  \\
    SC: unimodal    &  0.0 & 0.1 & 0.1  &  8.3 &  0.0 &  0.0 & 0.0  & 1.3  & 0.5  \\
    SC: crossmodal   &  1.8 & 11.1  & 3.88  &  64.5 &  32.7 & 21.7  & 16.8  &  46.0 & 32.1  \\ 
    SC: text  &  0.0 &  0.1 & 0.1  & 4.4  & 0.1  & 0.3  & 0.0  &  1.2 & 0.3  \\
    SC: video    &  28.8 &  21.8 & 6.5  & 75.1  & 34.3  & 29.3  &  36.0 & 55.3  & 41.0  \\
    QUAG-attention    &  28.9 & 22.3  & 6.0  &  74.4 &  35.0 & 27.3  & 37.3  & 54.1  & 41.1  \\
    \bottomrule
  \end{tabular}
\end{table}
\begin{table}
  \caption{Fine-grained performance of FrozenBiLM on MSRVTT-QA}
  \label{frozenbilm-msrvtt}
  \centering
  \begin{tabular}{lrrrrrr}
    \toprule
    Config & What & How & Color & Where & Who & When   \\
    \midrule
    Baseline    & 40.5  &  87.2 & 57.9  &  41.5 &  56.6 & 81.4   \\
    Lang-only    &  27.3 & 83.6  & 50.0  & 35.8  &  41.2 & 77.6    \\
    Vid-only    &  0.0 &  0.0 &  0.0 &  0.0 &  0.0 & 0.0   \\
    SC: unimodal    &  0.7 & 0.0  & 1.2  &  0.8 & 1.8  & 0.2    \\
    SC: crossmodal    &  27.1 &  83.4 &  50.9 & 32.9  & 41.1  & 66.3     \\
    SC: text    & 0.3  & 0.0  &  0.8 &  0.0 &  2.8 & 0.0    \\
    SC: video    &  39.8 &  85.5 &  58.8 &  41.9 &  55.4 & 80.9   \\
    QUAG-attention    &  39.9 &  86.2 &  58.1 & 42.7  & 55.2  & 81.1    \\
    \bottomrule
  \end{tabular}
\end{table}
\begin{table}[h!]
  \caption{Fine-grained performance of FrozenBiLM on NeXT-QA}
  \label{frozenbilm-nextqa}
  \centering
  \begin{tabular}{lrrr}
    \toprule
    Config & Causal & Temporal & Descriptive   \\
    \midrule
    Baseline    & 56.0  & 56.1  &  54.5     \\
    Lang-only    & 55.9  & 56.1  &  54.2     \\
    Vid-only    & 20.7  &  19.1 &  20.9    \\
    SC: unimodal    &  19.7 & 21.1  & 17.3     \\
    SC: crossmodal    &  56.1 & 56.5  & 54.3      \\
    SC: text     &  20.0 &  21.6 & 19.9    \\
    SC: video   &  56.1 & 56.1  & 54.5       \\
    QUAG-attention    & 55.9  & 55.8  & 54.1     \\
    \bottomrule
  \end{tabular}
\end{table}
\begin{table}[h!]
  \caption{Fine-grained performance of FrozenBiLM on ATH-Hard subset of NeXT-QA}
  \label{frozenbilm-nextqa-atph}
  \centering
  \begin{tabular}{lrrr}
    \toprule
    Config & Causal & Temporal    \\
    \midrule
    Baseline    &  55.2 & 56.3        \\
    Lang-only    & 55.5  & 56.2         \\
    Vid-only    &  20.0 & 20.1        \\
    SC: unimodal    & 20.7  & 22.5        \\
    SC: crossmodal    & 54.9  & 56.6         \\
    SC: text    & 20.2  &   22.3        \\
    SC: video    & 55.3  & 56.3      \\
    QUAG-attention    & 55.3  & 56.7       \\
    \bottomrule
  \end{tabular}
\end{table}
\subsection{Additional Results}
We evaluated QUAG on All-in-one model and find that, as the authors claim, the model utilizes both -- unimodal and cross-modal modality interactions. The results are summarized in Table \ref{table-quag-res-addn}.

\begin{table}[!ht]
  \caption{Short-circuit (SC)  results for All-in-one+ model on ActivityNet-QA (A-QA), and MSRVTT-QA (M-QA) datasets.}
  \label{table-quag-res-addn}
  \centering
        \centering
        \begin{tabular}{lrr}
        \toprule
        \multicolumn{1}{c}{} &
        \multicolumn{2}{c}{All-in-one+} \\
        \cmidrule{2-3}
             & ActivityNet-QA & MSRVTT-QA \\ 
            \midrule
            Acc & 41.9 & 43.1 \\ 
            text\_only & 23.5 & 20.8 \\
            vid\_only & 14.2 & 4.2 \\ 
            \midrule
            SC: unimodal & 11.4 & 3.8 \\ 
            SC: crossmodal & 20.6 & 27.6 \\ 
            SC: video & 19.2 & 12.7 \\ 
            SC: text & 5.6 & 7.3 \\ 
            \bottomrule
        \end{tabular}
    \end{table}
\section{CLAVI}
\subsection{Dataset Creation}
\label{sec:clavi_creation}
We curate CLAVI by leveraging Charades-STA (\url{https://prior.allenai.org/projects/data/charades/license.txt}) \citep{gao2017tall}, containing 9,848 videos of humans performing actions based on a short script written by composing predefined vocabulary that describe multiple daily actions. The videos are annotated with the start and end times of each action. The action category, the start, and the end of each action segment are referred to as the \emph{action tuple}.
Each video may contain more than two action tuples.
We select pairs of action tuples based on the uniqueness of the action category and complete exclusivity (that is no overlap between the occurrence of the actions).
In a given selected pair of action tuples, the two actions along with the inter-action region constitute the video segment. 
We ensure that the two action categories in the pair are distinct.
Additionally, to address temporal boundary ambiguities in the annotations, we filter out segments where either of the selected action classes occurs in close proximity to the segment boundaries 

We also extend the boundaries of the two actions in the pair. We define two boundary extensions -- out-extension and in-extension. The out-extension encompasses regions that are not a part of the selected segment but extend outwards in both directions into the original video. Similarly, in-extension extends inwards into the inter-action segment. To avoid temporal position bias \citep{hao2022can, otani2020challengesmr}, the lengths of the extension boundaries are selected randomly. However, since inter-action separation can affect their recognition \citep{bagad2023testoftime}, we constraint the inter-action separation in the original and the corresponding negative video  to be the same. That is, the sum of out-extension boundaries is always equal to the sum of in-extension boundaries.

We trim each boundary-extended contiguous segment from the original video to curate a positive video instance. To create the complementary video, we swap the boundary-extended action regions as shown in Figure \ref{fig:clavi}. Note that the region between the boundary-extended actions is unaffected. Swapping operation preserves the actions but only alters their chronology, and can be applied independently to question complements (unlike manipulations like video reversal \citep{wang2023paxion}). This independence provides fine-grained control to create a balanced benchmark for comprehensive analysis.

We create three types of questions using pre-defined templates and action-class annotations:

    1) \textbf{Existence (E) type}: The E-type questions for both the action classes follow the template \emph{"Was someone \textlangle A\textrangle?"}, where \emph{\textlangle A\textrangle} is one of two action classes in video. We use it as a positive control to verify if the model is able to correctly recognize the action classes. We use the exact same question for negative video instance as well, totalling to 4 control (questions, video, answer) instances for a Charades-extracted video segment.
    
    2) \textbf{Beginning/End (BE) type}: BE type questions the absolute location of the action in the video. The question is of the form, \emph{"Was the person {\textlangle A\textrangle} at the \{beginning/end\}?"} where \emph{\textlangle A\textrangle} is one of two action classes in the video, and we select one of \emph{beginning} and \emph{end}. Hence, for a given video and its negative, we have, in total, 8 instances of BE (questions, video, answer) tuples combined. Note that the answer for a given BE question is complemented in the negative video. 
    
    3) \textbf{Before/After (BA) type}: BA type comprises of questions on the relative order of occurrence of actions. The question is of the form \emph{"Did {\textlangle A1\textrangle} happen \{after/before\} \textlangle A2\textrangle?"}, where \emph{\textlangle A1\textrangle} and \emph{\textlangle A2\textrangle} are the selected action classes. We consider all the permutations of action classes. Hence, we have a total of 8 instances of BA type (questions, video, answer) tuples per extracted video. Similar to BE type, the answer is complemented in the negative video. 

Further, we add negative controls for E and BA type questions. A negative control action is an action that does not occur in the video. Since we want to probe only for temporal understanding, we keep the negative control action-class easy to detect by randomly selecting an action-class that does not contain any of the objects or actions in the original video. 
Hence, answering the negative control does not require understanding temporal cues in language and video and can be answered by object elimination. It serves the dual purpose of sanity check of learning and a baseline for learning by temporal shortcuts. The answer of negative control questions is always false. This adds two E type and sixteen BA type negative control questions for the video and its negative combined. Hence, including the negative control questions, each video in CLAVI is associated with 19 questions: 2 E, 4 BE, 4 BA, 1 E negative control and 8 BA negative controls. The ratio of "yes":"no" answers is 6:13. 
\subsection{Comparison with Existing Datasets}
\label{sec: clavi_dataset_comp}
We provide a comparison of size of CLAVI with established VideoQA datasets in Table \ref{tab: clavi_comp}. 


\begin{table}
    \caption{Comparison of CLAVI with other other VideoQA datasets sorted in the reverse order of recency.}

    \label{tab: clavi_comp}
    \centering
    \centering
    \begin{tabular}{l c}
    \toprule
        Dataset & Number of (V,Q,A) samples \\ 
        \midrule
        MSRVTT-QA \citep{xu2017video} & 243K  \\ 
        ActivityNet-QA \citep{yu2019activitynet} & 58K  \\ 
        Social-IQ QA \citep{zadeh2019social} & 7.5K  \\ 
        NeXT-QA \citep{xiao2021next} & 52K \\ 
        iVQA \citep{yang2021justask} & 10K \\ 
        STAR \citep{wu2021star} & 60K\\
        EgoTaskQA \citep{jia2022egotaskqa} & 40K \\
        FIBER \citep{castro-etal-2022-fiber} & 28K \\
        NewsQA \citep{jahagirdar2023watching} & 8.6K \\
        \textbf{CLAVI (Ours)} & \textbf{114K}  \\
        \bottomrule
    \end{tabular}
\end{table}

\subsection{Comprehensive List of Questions}
\label{sec:list_questions}
We provide a comprehensive list of the questions for the example presented in Fig 2 of the main paper.
We define the actions as:
\textbf{A}: \textit{holding clothes}
\textbf{B}: \textit{taking food}
\textbf{C}: \textit{washing mirror},
where action A occurs before action B in the original video and action C does not occur anywhere in the original video.

Enlisted below are the questions and its complement (Q and Q' respectively) for the video (V) (that is event A occurs after event B). Note that the color of the panel is representative of the answer of the question (red: ``no'', green: ``yes'').

\textbf{E-Type}:\\
\fcolorbox{green}{green!20}{%
    \parbox{\textwidth}{%
    \textbf{Q\phantom{'}:} Was someone holding clothes?
    }%
}
\fcolorbox{green}{green!20}{%
    \parbox{\textwidth}{%
    \textbf{Q\phantom{'}:} Was someone taking food?
    }%
}

\textbf{E-Type (negative control)}:\\
\fcolorbox{red}{red!20}{%
    \parbox{\textwidth}{%
    \textbf{Q\phantom{'}:} Was someone washing mirror?
    }%
}

\textbf{BE-Type}\\
\fcolorbox{green}{green!20}{%
    \parbox{\textwidth}{%
    \textbf{Q\phantom{'}:} Was the person holding clothes at the \textbf{beginning}?
    }%
}
\fcolorbox{red}{red!20}{%
    \parbox{\textwidth}{%
    \textbf{Q':} Was the person holding clothes at the \textbf{end}?
    }%
}

\fcolorbox{green}{green!20}{%
    \parbox{\textwidth}{%
    \textbf{Q\phantom{'}:} Was the person taking food at the \textbf{end}?
    }%
}
\fcolorbox{red}{red!20}{%
    \parbox{\textwidth}{%
    \textbf{Q':} Was the person taking food at the \textbf{beginning}?
    }%
}

\textbf{BA-Type}\\
\fcolorbox{green}{green!20}{%
    \parbox{\textwidth}{%
    \textbf{Q\phantom{'}:} Did holding clothes happen \textbf{before} taking food?
    }%
}
\fcolorbox{red}{red!20}{%
    \parbox{\textwidth}{%
    \textbf{Q':} Did holding clothes happen \textbf{after} taking food?
    }%
}

\fcolorbox{green}{green!20}{%
    \parbox{\textwidth}{%
    \textbf{Q\phantom{'}:} Did taking food happen \textbf{after} holding clothes?
    }%
}
\fcolorbox{red}{red!20}{%
    \parbox{\textwidth}{%
    \textbf{Q':} Did taking food happen \textbf{before} holding clothes?
    }%
}

\textbf{BA-Type (negative-control)}\\
\fcolorbox{red}{red!20}{%
    \parbox{\textwidth}{%
    \textbf{Q':} Did washing mirror happen \textbf{before} holding clothes?
    }%
}
\fcolorbox{red}{red!20}{%
    \parbox{\textwidth}{%
    \textbf{Q':} Did washing mirror happen \textbf{after} holding clothes?
    }%
}

\fcolorbox{red}{red!20}{%
    \parbox{\textwidth}{%
    \textbf{Q':} Did holding clothes happen \textbf{before} washing mirror?
    }%
}
\fcolorbox{red}{red!20}{%
    \parbox{\textwidth}{%
    \textbf{Q':} Did holding clothes happen \textbf{after} washing mirror?
    }%
}

\fcolorbox{red}{red!20}{%
    \parbox{\textwidth}{%
    \textbf{Q':} Did washing mirror happen \textbf{before} taking food?
    }%
}
\fcolorbox{red}{red!20}{%
    \parbox{\textwidth}{%
    \textbf{Q':} Did washing mirror happen \textbf{after} taking food?
    }%
}

\fcolorbox{red}{red!20}{%
    \parbox{\textwidth}{%
    \textbf{Q':} Did taking food happen \textbf{before} washing mirror?
    }%
}
\fcolorbox{red}{red!20}{%
    \parbox{\textwidth}{%
    \textbf{Q':} Did taking food happen \textbf{after} washing mirror?
    }%
}


Enlisted below are the questions and its negatives (Q and Q' respectively) for the negative video instance (V') (that is event B occurs after event A). 

\textbf{E-Type}:\\
\fcolorbox{green}{green!20}{%
    \parbox{\textwidth}{%
    \textbf{Q\phantom{'}:} Was someone holding clothes?
    }%
}
\fcolorbox{green}{green!20}{%
    \parbox{\textwidth}{%
    \textbf{Q\phantom{'}:} Was someone taking food?
    }%
}

\textbf{E-Type (negative control)}:\\
\fcolorbox{red}{red!20}{%
    \parbox{\textwidth}{%
    \textbf{Q\phantom{'}:} Was someone washing mirror?
    }%
}

\textbf{BE-Type}\\
\fcolorbox{red}{red!20}{%
    \parbox{\textwidth}{%
    \textbf{Q\phantom{'}:} Was the person holding clothes at the \textbf{beginning}?
    }%
}
\fcolorbox{green}{green!20}{%
    \parbox{\textwidth}{%
    \textbf{Q':} Was the person holding clothes at the \textbf{end}?
    }%
}

\fcolorbox{red}{red!20}{%
    \parbox{\textwidth}{%
    \textbf{Q\phantom{'}:} Was the person taking food at the \textbf{end}?
    }%
}
\fcolorbox{green}{green!20}{%
    \parbox{\textwidth}{%
    \textbf{Q':} Was the person taking food at the \textbf{beginning}?
    }%
}

\textbf{BA-Type}\\
\fcolorbox{red}{red!20}{%
    \parbox{\textwidth}{%
    \textbf{Q\phantom{'}:} Did holding clothes happen \textbf{before} taking food?
    }%
}
\fcolorbox{green}{green!20}{%
    \parbox{\textwidth}{%
    \textbf{Q':} Did holding clothes happen \textbf{after} taking food?
    }%
}

\fcolorbox{red}{red!20}{%
    \parbox{\textwidth}{%
    \textbf{Q\phantom{'}:} Did taking food happen \textbf{after} holding clothes?
    }%
}
\fcolorbox{green}{green!20}{%
    \parbox{\textwidth}{%
    \textbf{Q':} Did taking food happen \textbf{before} holding clothes?
    }%
}

\textbf{BA-Type (negative-control)}\\
\fcolorbox{red}{red!20}{%
    \parbox{\textwidth}{%
    \textbf{Q':} Did washing mirror happen \textbf{before} holding clothes?
    }%
}
\fcolorbox{red}{red!20}{%
    \parbox{\textwidth}{%
    \textbf{Q':} Did washing mirror happen \textbf{after} holding clothes?
    }%
}

\fcolorbox{red}{red!20}{%
    \parbox{\textwidth}{%
    \textbf{Q':} Did holding clothes happen \textbf{before} washing mirror?
    }%
}
\fcolorbox{red}{red!20}{%
    \parbox{\textwidth}{%
    \textbf{Q':} Did holding clothes happen \textbf{after} washing mirror?
    }%
}

\fcolorbox{red}{red!20}{%
    \parbox{\textwidth}{%
    \textbf{Q':} Did washing mirror happen \textbf{before} taking food?
    }%
}
\fcolorbox{red}{red!20}{%
    \parbox{\textwidth}{%
    \textbf{Q':} Did washing mirror happen \textbf{after} taking food?
    }%
}

\fcolorbox{red}{red!20}{%
    \parbox{\textwidth}{%
    \textbf{Q':} Did taking food happen \textbf{before} washing mirror?
    }%
}
\fcolorbox{red}{red!20}{%
    \parbox{\textwidth}{%
    \textbf{Q':} Did taking food happen \textbf{after} washing mirror?
    }%
}
\subsection{Dataset Metrics}
\begin{figure}
  \centering
  \includegraphics[width=\textwidth]{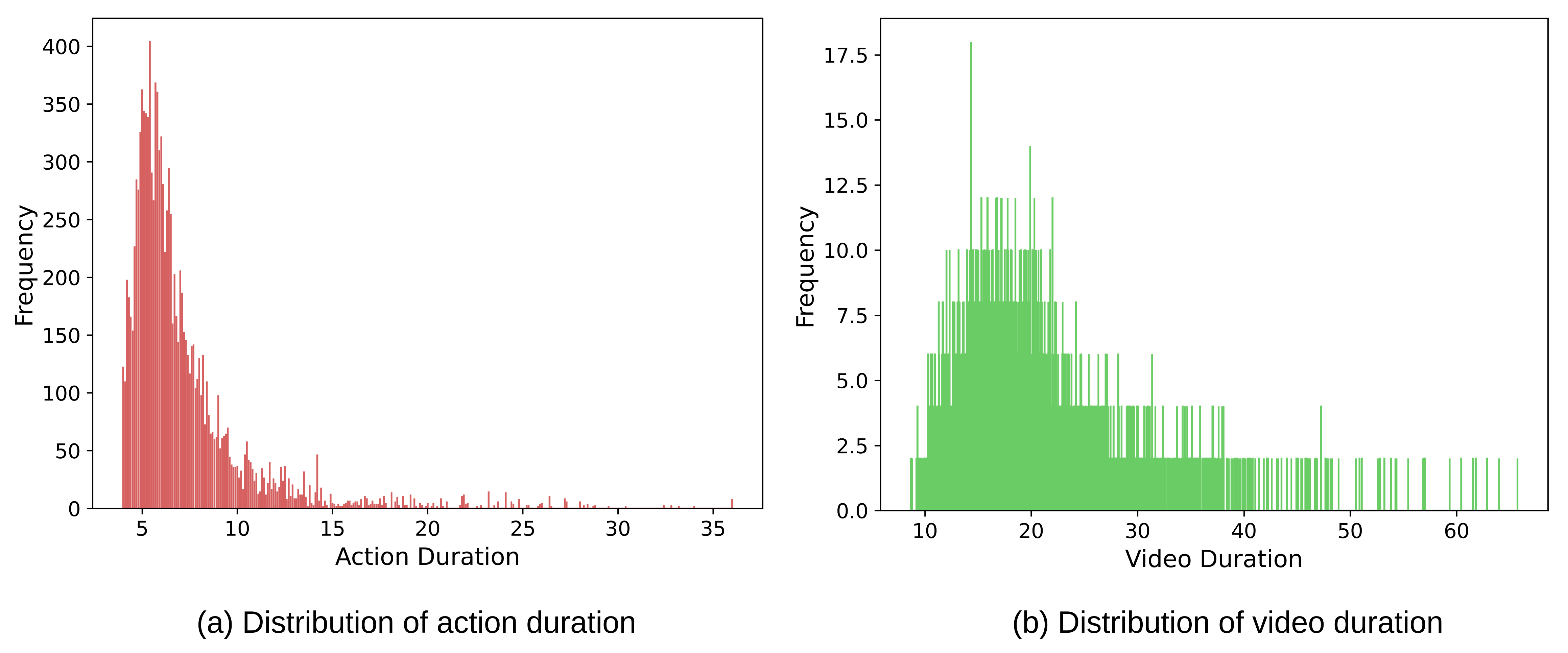}
  \caption{Distribution of length of (a) action and (b) video durations}
  \label{fig:metrics_2}
\end{figure}

The duration of individual action in CLAVI lies in the range [4.0 sec, 36.0 sec]; the average length of action is \textbf{7.7 $\mathbf{\pm}$ 3.42} sec. The average video length is \textbf{19.95 $\mathbf{\pm}$ 7.34} secs and the range is [8.67 sec, 65.73 sec]. We plot the distribution of the action and video durations in Fig. \ref{fig:metrics_2}.

\begin{figure}
  \centering
  \includegraphics[width=\textwidth]{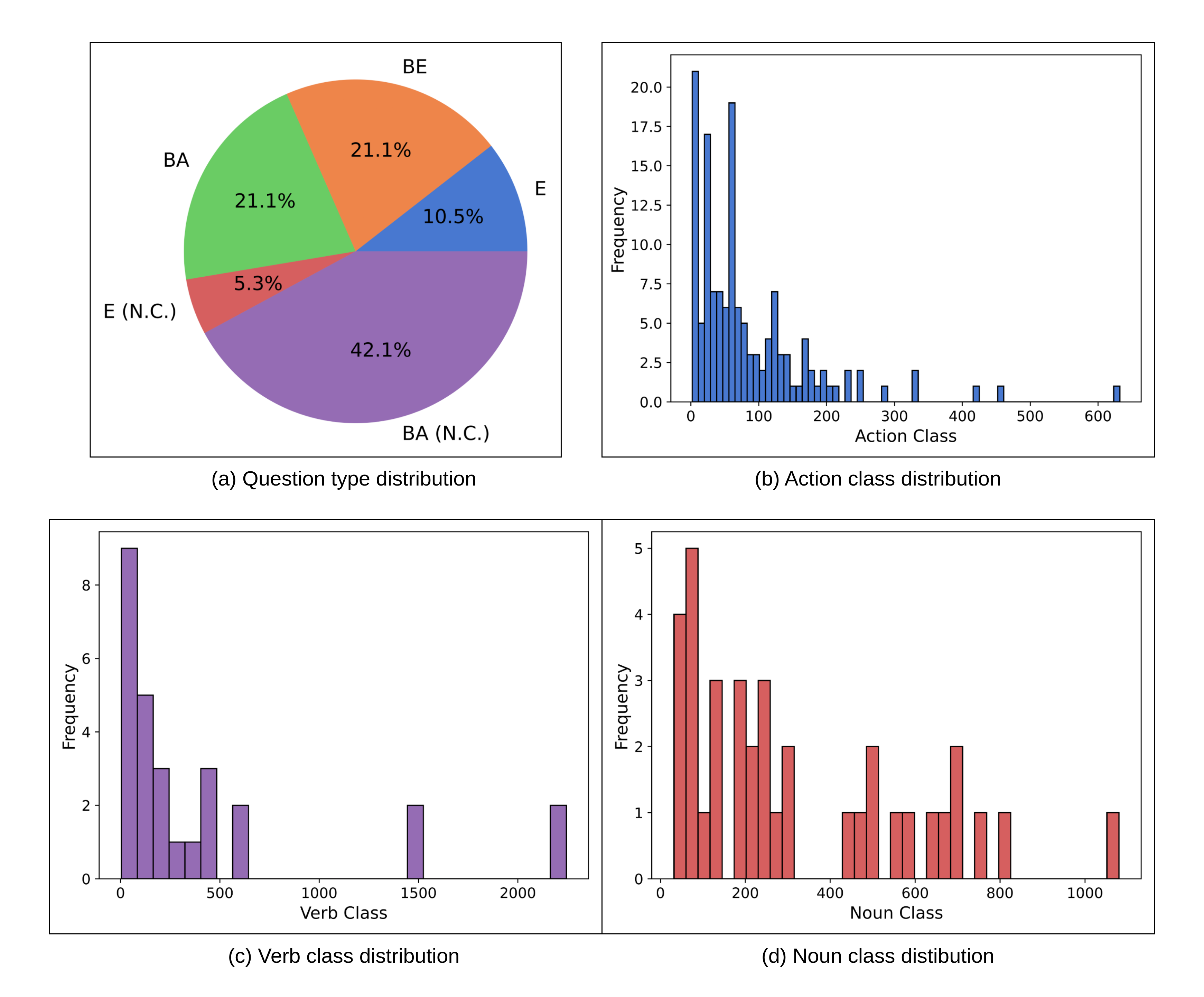}
  \caption{Metrics of the dataset (a) distribution of question types (same for training and testing set), (b) histogram plot of frequencies of action classes (c) histogram plot of frequencies of verb classes (d) histogram plot of frequencies of noun classes.}
  \label{fig:metrics_mega}
\end{figure}

CLAVI consists of \textbf{141} unique action classes. Each action class is composed of noun (objects) and verb. There are \textbf{37} unique noun classes and \textbf{28} unique verb classes. We show the frequency distributions of action, verb and noun classes in Fig \ref{fig:metrics_mega}.

\subsection{Experiment Details}
\label{sec:clavi_exp_details_supp}
As mentioned in the main manuscript, we use the official checkpoints, finetuning code and hyper-parameters of JustAsk \href{https://github.com/antoyang/just-ask}{[website]},  FrozenBiLM \href{https://github.com/antoyang/FrozenBiLM}{[website]} , Singularity-Temporal \href{https://github.com/jayleicn/singularity}{[website]}, and All-in-one+ \href{https://github.com/showlab/all-in-one}{[website]}. For JustAsk, we use the checkpoint of the model pretrained on HowToVQA69M and WebVidVQA3M. For FrozenBiLM, we use the WebVid10M-pretrained checkpoint. All-in-one+ is pretrained on eight datasets comprising of both images and videos (videos: Webvid, YT-Temporal-180M, HowTo100M and images: CC3M, CC12M, COCO, Visual Genome, SBU Captions). Singularity-Temporal is pretrained on a 17.28M images and video subset (images: COCO, Visual Genome, SBU Captions, CC3M, CC12M and videos: WebVid). We have depicted the finetuning details in Table \ref{table:clavi_ft_det}.

\begin{table}
    \caption{Hyperparameters and checkpoint details of CLAVI finetuning experiment}
    \centering
    \footnotesize 
    \begin{tabularx}{\textwidth}{l X l l}
    \toprule
        Model &  Checkpoint & Epochs & LR \\ 
        \midrule
        JustAsk &  HowToVQA69M, WebVidVQA3M & 20 & 1.00E-05 \\ 
        FrozenBiLM & WebVid10M & 20 & 5.00E-05 \\ 
        All-In-One+ & Webvid, YT-Temporal-180M, HowTo100M, CC3M, CC12M, COCO, Visual Genome, SBU Captions & 10 & 1.00E-04 \\ 
        Singularity-T & COCO, Visual Genome, SBU Captions, CC3M, CC12M, WebVid & 20 & 1.00E-05 \\ 
        \bottomrule
    \end{tabularx}
  \label{table:clavi_ft_det}
\end{table}
\subsection{Fine-grained Accuracies}
In Table \ref{clavi-fine-acc-q} 
we provide error bars for the finetuning experiments. The experiments were performed thrice on the same hardware with the same set of hyperparameters.

\begin{table}
  \caption{Fine-grained performance (\% of accuracy) on CLAVI for question (Q) and complement question (Q'), video (V) and complement video (V') (Note: N.C. refers to Negative Control)}
  \label{clavi-fine-acc-q}
  \centering
  \begin{tabular}{|l|llcccc}
    \toprule
    V/V' & Question & Q/Q' & JustAsk & FrozenBiLM & Singularity-T & All-in-one+   \\
    \midrule
    \multirow{7}{*}{V} & E-type & Q &  $89.55 \pm 0.01$  & $87.51 \pm 0.00$ & $90.75 \pm 0.03$ &  $86.08 \pm 2.59$\\
    \cline{2-7}
     & E-type (N.C.) & - &  $75.28 \pm 0.02$  & $88.66 \pm 0.00$ & $79.16 \pm 0.03$  & $69.34 \pm 11.72$ \\
     \cline{2-7}
     & \multirow{2}{*}{BE-type} & Q &   $69.80 \pm 0.07$ & $69.15 \pm 0.01$ & $98.23 \pm 0.01$ & $99.31 \pm 0.84$\\
     &  & Q' &  $30.58 \pm 0.07$ &  $73.25 \pm 0.01$ & $1.87 \pm 0.01$ & $0.73 \pm 0.84$\\
     \cline{2-7}
     & \multirow{2}{*}{BA-type} & Q &  $27.81 \pm 0.02$  & $56.88 \pm 0.01$ & $62.55 \pm 0.09$ & $25.82 \pm 5.49$\\
     &  & Q' &  $72.31 \pm 0.02$  &  $86.79 \pm 0.01$ & $37.23 \pm 0.09$ & $74.31 \pm 0.84$ \\
     \cline{2-7}
     & \multirow{1}{*}{BA-type (N.C.)} & - &   $98.23 \pm 0.00$ &  $96.79 \pm 0.00$ & $ 93.72\pm 0.03$ & $ 98.44\pm 1.02$\\
     \midrule
         \multirow{7}{*}{V'} & E-type & Q &  $89.17 \pm 0.01$  &   $86.96 \pm 0.01$ & $90.58 \pm 0.02$ & $86.03 \pm 2.66$ \\
    \cline{2-7}
     & E-type (N.C.) & Q &  $76.10 \pm 0.03$  & $88.45 \pm 0.01$  & $79.04 \pm 0.03$ & $ 69.17\pm 11.26$\\
     \cline{2-7}
    & \multirow{2}{*}{BE-type} & Q &  $30.18 \pm 0.07$  &  $73.61 \pm 0.01$  & $1.80 \pm 0.01$ & $ 0.76\pm 1.00$\\
     &  & Q' &  $69.88 \pm 0.07$  & $70.00 \pm 0.02$  &  $98.28 \pm 0.01$ & $ 99.12\pm 1.02$\\
     \cline{2-7}
     & \multirow{2}{*}{BA-type} & Q &   $71.61 \pm 0.02$ &  $85.43 \pm 0.01$  & $38.00 \pm 0.08$ & $ 74.24\pm 5.12$\\
     &  & Q' &  $28.34 \pm 0.02$  & $54.44 \pm 0.00$ &  $62.15 \pm 0.07$ & $ 25.90\pm 4.93$ \\
     \cline{2-7}
     & \multirow{1}{*}{BA-type (N.C.)} & - &  $98.51 \pm 0.00$  &   $96.87 \pm 0.00$ & $93.51 \pm 0.03$ & $ 98.46\pm 1.04$ \\
    \bottomrule
  \end{tabular}
\end{table}

\subsection{Representation Sensitivity Analysis}
\label{sec:clavi_additional_analysis}
CLAVI can be used for diverse analyses to understand and interpret the joint multimodal representations in VideoQA models. We present one such analysis here. We want to find out the difference in representations between correctly and wrongly-answered complement pairs. Ideally, the complement pairs should have distinctly dissimilar representations to be answered correctly. 

We use L2 norm as the distance metric. For CLAVI, we construct complements by augmenting the sequence of the frames (video complements) or replacing before/after and beginning/end (text complement). Hence, we cannot directly compute the distance between the attention matrices of the complements because they contain different tokens (text complement) or different order of the same tokens (video complement). 
We solve this by finding token correspondence between the complement pairs for each layer and head.
By treating each attention matrix as a graph, we model the matrix alignment problem to finding the node correspondence between two isomorphic weighted directed complete graphs. Node correspondence between two graphs can be viewed as an instance of a linear sum assignment problem. That is, we want to learn a permutation transformation so that the two attention matrices as similar. We define similarity as negative of L2 distance. We solve this using modified Jonker-Volgenant algorithm as described by \cite{crouse2016implementing}. 

\begin{figure}
  \centering
  \includegraphics[width=\textwidth]{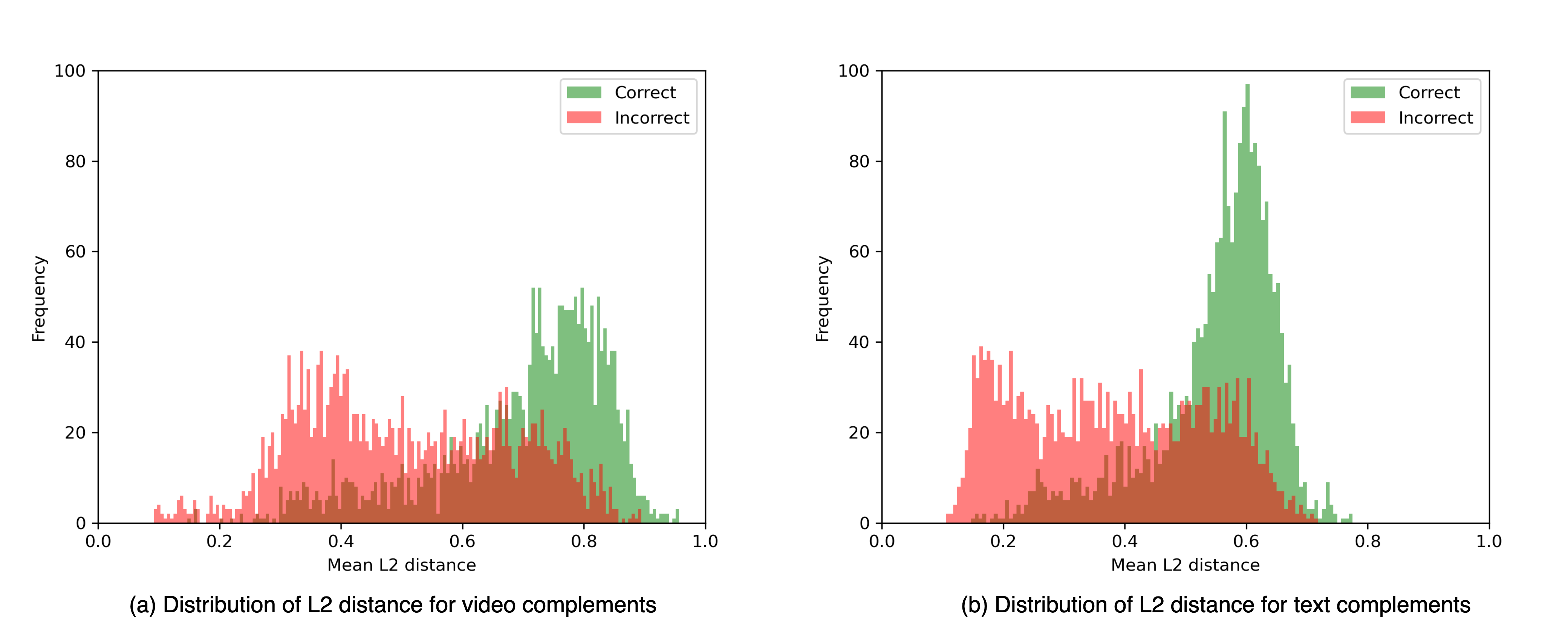}
  \caption{Histogram plots of mean l2 distance between complement BA-type pairs for (a) video and (b) text complements for FrozenBiLM predictions (note that green is consistently correct; that is both the pairs in the complements are correctly answered, Similarly, red is consistently incorrect; that is at least one of the instance from the complement pair is incorrectly answered). }
  \label{fig:iclr_clavi_addn_anal}
\end{figure}

We plot the histogram of L2 distance (averaged over heads and layers) for BA-type video and question complements in Figure \ref{fig:iclr_clavi_addn_anal}. As expected, we find that if the answer is correct, then the average L2 distance is generally higher (skewed towards right). The mean and variance values L2 mean distribution of the correctly and incorrectly answered complement pairs is summarized in Table \ref{tab: res_clavi_repr_anal}. We find that the correct predictions have higher mean and lower variance than the incorrectly inferred complement pairs. These findings validate that the model in indeed learning joint multimodal representations rather than creating its illusion.

\begin{table}
    \centering
    \caption{Statistics of L2 distance values between aligned attention matrices of BA-type CLAVI questions, averaged over all heads and layers for FrozenBiLM. We report the statistics separately for correctly and incorrectly answered consistent complement predictions.}
    \label{tab: res_clavi_repr_anal}
    \begin{tabular}{l l l l }
    \toprule
        Type & Consistent Prediction & Mean & Variance \\ 
        \midrule
        \multirow{2}{*}{Video complement} & Correct & 0.70 & 0.02 \\ 
        \cline{2-4}
        ~ & Incorrect & 0.50 & 0.03 \\
        \midrule
        \multirow{2}{*}{Text complement} & Correct & 0.55 & 0.01 \\
        \cline{2-4}
        ~ & Incorrect & 0.38 & 0.02 \\ 
    \bottomrule
    \end{tabular}
\end{table}

\section{Limitations and Future Work}
\label{future_work}
Our dataset is intentionally simple, so as to focus the benchmark only on simple temporal sequence understanding, which preempts spatio-temporal referential understanding. We plan to include  more complex temporal organizations of action classes like containment and partial-overlap that are defined using prepositions like \textit{during} and \textit{while} in future work. As the current state-of-the-art models catch-up to our benchmark, our future plan is to curate a more complex dataset with more natural questions that include temporal referring expressions with similar balanced doubly-negative strategy. 

\subsection{Datasheet} 
\label{sec:datasheet}
In this section we provide a more detailed documentation of the dataset with the intended uses. We
base ourselves on the datasheet proposed by \citet{gebru2021datasheets}

\subsubsection{Motivation}
\begin{itemize}

\item \textbf{For what purpose was the dataset created?} CLAVI is curated to diagnose and stress-test the joint multimodal understanding in VideoQA models. It uses temporal complements in video and question domains to assess if the models are able to learn the temporal structure from both the modalities.

\item \textbf{Who created the dataset and on behalf of which entity?} 
CLAVI was curated by a team of researchers from the Agency For Science, Technology and Research (A*STAR), Singapore, comprising of Ishaan Singh Rawal, Alexander Matyasko, Shantanu Jaiswal, Basura Fernando and Cheston Tan.

\item \textbf{Who funded the creation of the dataset?} 
Agency For Science Technology and Research (A*STAR), Singapore

\end{itemize}
\subsubsection{Composition}
\begin{itemize}
    
\item \textbf{What do the instances that comprise the dataset represent?} Each instance in CLAVI comprises of a video\_id (from Charades-STA (freely available online)), question, question type, and answer (``yes'' or ``no''). Due to licensing issues, we do not release the videos but have provided the scripts for curating the dataset from Charades-STA.

\item \textbf{How many instances are there in total?}
CLAVI consists of 6,018 videos composing of 
3,830 training and 2,188 testing videos. Each video is associated with 19 question-answer pairs, hence 114,342 data-points (72,770 training and 41,572 testing).

\item \textbf{Does the dataset contain all possible instances or is it a sample (not necessarily random) of
instances from a larger set?}
The videos and question answer pairs in CLAVI are generated by manipulating real-world real-world videos from Charades-STA. In theory, we
can generate more instances with more  videos with temporal annotations. We will release the code to generate the true and counterfactual video and question instances.

\item \textbf{What data does each instance consist of?} Each instance in CLAVI comprises of a video\_id (from Charades-STA), question, question type, and answer (``yes'' or ``no'').

\item \textbf{Is there a label or target associated with each instance?} Yes, each question is associated with a ``type'' label depending on the type of the question (types described in the main manuscript).

\item \textbf{Is any information missing from individual instances?} No, all the instances have 
complete information the corresponding attributes.

\item \textbf{Are relationships between individual instances made explicit?}
Yes, the \textit{video\_name} attribute if of the form XXXXXXXX\_1 for the original video segment and XXXXXXXX\_2 for the counterfactual video segment, where XXXXXXXX is a unique 8-digit video id. The relationship between counterfactual questions is tabulated in the README file of the dataset.

\item \textbf{Are there recommended data splits (e.g., training, development/validation, testing)?}
We provide the split files which are curated from the original split files of Charades.

\item \textbf{Are there any errors, sources of noise, or redundancies in the dataset?}
No. The owners of Charades do not report any known errors. And since our data is generated by machine, we do not expect any errors. For unforeseen errors in temporal annotation boundaries in the original dataset, we eliminate it by selecting the segments where the actions of interest do not occur in the immediate neighbourhood (detailed in the main manuscript).

\item \textbf{Is the dataset self-contained, or does it link to or otherwise rely on external resources?}
This dataset provides video IDs from the Charades dataset under their Non-Commercial
license.

\item \textbf{Does the dataset contain data that might be considered confidential? }
No. We curate our dataset from publicly and non-commercially available Charades dataset.

\item \textbf{Does the dataset identify any sub-populations (e.g., by age, gender)?}
No. While it is possible to identify gender from Charades temporal captions, we do not use it in the curation of CLAVI. We only use neutral pronoun \emph{someone}.

\item \textbf{Is it possible to identify individuals (i.e., one or more natural persons), either directly or
indirectly (i.e., in combination with other data) from the dataset? }
No. The owners of Charades dataset have anonymized the subject information.

\item \textbf{Does the dataset contain data that might be considered sensitive in any way?}
No. The owners of Charades dataset ensure this and we curate CLAVI from Charades.

\end{itemize}
\subsubsection{Collection Process}
\begin{itemize}

\item \textbf{How was the data associated with each instance acquired?}
Each sample of CLAVI associates with a question, answer (yes/no) and video-id from Charades dataset.

\item \textbf{What mechanisms or procedures were used to collect the data (e.g., hardware apparatuses
or sensors, manual human curation, software programs, software APIs)? } We design a template-based VideoQA generation process to generate data each instance from Charades.

\item \textbf{Who was involved in the data collection process (e.g., students, crowdworkers, contractors)
and how were they compensated (e.g., how much were crowdworkers paid)? } Not applicable

\item \textbf{If the dataset is a sample from a larger set, what was the sampling strategy (e.g.,
deterministic, probabilistic with specific sampling probabilities)?}
Yes. We have outlined the process of filtering the data in detail in the appendix.

\item \textbf{Who was involved in the data collection process (e.g., students, crowdworkers, contractors) and how were they compensated (e.g., how much were crowdworkers paid)?}
Not applicable.

\item \textbf{Over what timeframe was the data collected? }
Our dataset is generated from Charades. We generate the dataset from
February 2023 to June 2023.

\item \textbf{Were any ethical review processes conducted (e.g., by an institutional review board)?}
Not Applicable.

\item \textbf{Does the dataset relate to people?} Yes. The Charades dataset contains videos of humans performing actions and we use it to curate CLAVI under their non-commercial license. However, we do not use the information pertaining to humans anyway.

\item \textbf{Did you collect the data from the individuals in question directly, or obtain it via third
parties or other sources (e.g., websites)?} No. Our video data is from Charades under their Non-Commercial license. Charades
Homepage: https://prior.allenai.org/projects/charades.

\item \textbf{Were the individuals in question notified about the data collection? } Not applicable. We curate our dataset from Charades and the original owners have ensured this.

\item \textbf{Did the individuals in question consent to the collection and use of their data?} Not applicable. We curate our dataset from Charades and the original owners have ensured this.

\item \textbf{If consent was obtained, were the consenting individuals provided with a mechanism to
revoke their consent in the future or for certain uses?} Not applicable.
\end{itemize}
\subsubsection{Preprocessing/cleaning/labeling}
\begin{itemize}

\item \textbf{Was any preprocessing/cleaning/labeling of the data done (e.g., discretization or bucketing,
tokenization, part-of-speech tagging, SIFT feature extraction, removal of instances, processing
of missing values)} No.

\item \textbf{Was the “raw” data saved in addition to the preprocessed/cleaned/labeled data (e.g., to support unanticipated future uses)? } No. The raw data (Charades) is distributed under Non-Commerical license.

\item \textbf{Is the software that was used to preprocess/clean/label the data available?}
We automatically curated the dataset using Python3 and shell-scripting, both of which are acessible and widely used.
\end{itemize}
\subsubsection{Uses}
\begin{itemize}

\item \textbf{Has the dataset been used for any tasks already?} Video Question Answering.

\item \textbf{Is there a repository that links to any or all papers or systems that use the dataset?} No.

\item \textbf{What (other) tasks could the dataset be used for? } Video-Text and Text-Video retrieval.

\item \textbf{Is there anything about the composition of the dataset or the way it was collected and preprocessed/cleaned/labeled that might impact future uses? } No.

\item \textbf{Are there tasks for which the dataset should not be used?} No, we do not foresee any such usage as of now.

\item \textbf{Will the dataset be distributed to third parties outside of the entity (e.g., company, institution,
organization) on behalf of which the dataset was created? }
CLAVI is an academic dataset for public non-commercial use.

\item \textbf{How will the dataset be distributed (e.g., tarball on website, API, GitHub)?}
The scripts to curate the dataset will be released on GitHub.

\item \textbf{When will the dataset be distributed?}
Latest by the official paper acceptance.

\item \textbf{Will the dataset be distributed under a copyright or other intellectual property (IP) license,
and/or under applicable terms of use (ToU)? }
Yes. The dataset will be released under GPL3.0 license and terms of usage will be outlined on the dataset hosting website along with the license and the required scripts. 

\textbf{Have any third parties imposed IP-based or other restrictions on the data associated with the
instances?}
No.

\textbf{Do any export controls or other regulatory restrictions apply to the dataset or to individual
instances?}
No.
\end{itemize}

\subsubsection{Maintenance}
\begin{itemize}

\item  \textbf{Who will be supporting/hosting/maintaining the dataset?}
Ishaan Singh Rawal and Cheston Tan

\textbf{How can the owner/curator/manager of the dataset be contacted (e.g., email address)?}
The owners can be contacted via email: \{rawal\_ishaan\_singh, cheston\_tan\}@cfar.a-star.edu.sg

\item  \textbf{Is there an erratum?}
Not yet. Errata, if any, will be communicated through Github.

\item  \textbf{Will the dataset be updated (e.g., to correct labeling errors, add new instances, delete instances)?}
Updates, if any, will be clearly mentioned on GitHub.

\item  \textbf{If the dataset relates to people, are there applicable limits on the retention of the data associated
with the instances (e.g., were the individuals in question told that their data would be retained
for a fixed period of time and then deleted)?}
No.

\item  \textbf{Will older versions of the dataset continue to be supported/hosted/maintained?}
There are no older versions of the datasets at the current moment. However, we plan to appropriately version the curation scripts to ensure reproducability. 

\item  \textbf{If others want to extend/augment/build on/contribute to the dataset, is there a mechanism for
them to do so? }
Yes, we will provide the necessary code files with the dataset. We will host the scripts to curate the datasets on Github, welcoming open-source contributions.
\end{itemize}


\end{document}